\documentclass[sigconf]{acmart}

\usepackage{subcaption}
\usepackage{algorithm}
\usepackage{algorithmic}
\usepackage{balance}

\AtBeginDocument{%
  \providecommand\BibTeX{{%
    \normalfont B\kern-0.5em{\scshape i\kern-0.25em b}\kern-0.8em\TeX}}}

\setcopyright{acmlicensed}
\copyrightyear{2024}
\acmYear{2024}
\acmDOI{XXXXXXX}

\acmConference[MM'24]{Make sure to enter the correct
  conference title from your rights confirmation email}{October 28 - November 1,
  2024}{Melbourne, Australia.}
\acmISBN{978-1-4503-XXXX-X/24/111}

\begin{document}

\title[Uncovering Capabilities of Model Pruning in Graph Contrastive Learning]{Uncovering Capabilities of Model Pruning \\ in Graph Contrastive Learning}


\author{Junran Wu}
\orcid{0000-0001-6742-4332}
\affiliation{%
  \institution{State Key Laboratory of Complex \& Critical Software Environment, Beihang University}
  \city{Beijing}
  \country{China}}
\email{wu_junran@buaa.edu.cn}

\author{Xueyuan Chen}
\orcid{0000-0002-1660-5325}
\authornote{Equal Contribution.}
\affiliation{%
  \institution{State Key Laboratory of Complex \& Critical Software Environment, Beihang University}
  \city{Beijing}
  \country{China}}
\email{xueyuanchen@buaa.edu.cn}

\author{Shangzhe Li}
\orcid{0000-0001-6742-4332}
\authornote{Corresponding Author.}
\affiliation{%
  \institution{School of Statistics and Mathematics, Central University of Finance and Economics}
  \city{Beijing}
  \country{China}}
\email{shangzheli@cufe.edu.cn}
\renewcommand{\shortauthors}{Junran Wu, Xueyuan Chen and Shangzhe Li.}

\begin{abstract}
Graph contrastive learning has achieved great success in pre-training graph neural networks without ground-truth labels. 
Leading graph contrastive learning follows the classical scheme of contrastive learning, forcing model to identify the essential information from augmented views.
However, general augmented views are produced via random corruption or learning, which inevitably leads to semantics alteration. 
Although domain knowledge guided augmentations alleviate this issue, the generated views are domain specific and undermine the generalization.
In this work, motivated by the firm representation ability of sparse model from pruning, we reformulate the problem of graph contrastive learning via contrasting different model versions rather than augmented views.
We first theoretically reveal the superiority of model pruning in contrast to data augmentations.
In practice, we take original graph as input and dynamically generate a perturbed graph encoder to contrast with the original encoder by pruning its transformation weights. 
Furthermore, considering the integrity of node embedding in our method, we are capable of developing a local contrastive loss to tackle the hard negative samples that disturb the model training.
We extensively validate our method on various benchmarks regarding graph classification via unsupervised and transfer learning. Compared to the state-of-the-art (SOTA) works, better performance can always be obtained by the proposed method.
\end{abstract}





\maketitle

\section{Introduction}
In light of the depletion of labeled data and the hardness of annotation, plenty of research attention has been moved from supervised learning to unsupervised learning~\cite{he2020momentum,wu2023sega,zhu2023uaed,zhu2024do}. While in graph domain, the same issue exists~\cite{hu2020strategies}. 
Correspondingly, referring to the design for unlabeled data training in natural language processing~\cite{yang2019xlnet,zhu2024hill} and computer vision~\cite{he2020momentum}, several solutions have been dug out through plenty of research efforts and collectively called graph contrastive learning, such as GraphCL~\cite{you2020graph} and AD-GCL~\cite{suresh2021adversarial}.

In general, graph contrastive learning sticks to the twin-tower architecture of contrastive learning~\cite{wu2018unsupervised,chen2020simple}, in which two augmented views are generated from the input graph, and the model loss (i.e., NT-Xent loss~\cite{chen2020simple}) maximizes the mutual information between the two output embeddings of the two augmented views.
With this design, the trained model is capable of capturing the essential information of graphs \cite{ma2021deep,wu2022simple,zhu2023hitin}.
Moreover, researchers have found that the quality of views influences the performance of contrastive learning models~\cite{tian2020makes}. Therefore, plenty of research efforts have been devoted to the generation of effective views that lead to better performance for graphs through the data augmentation~\cite{suresh2021adversarial,you2021graph}.

\begin{figure*}[ht]
\centering
\includegraphics[width=0.8\textwidth]{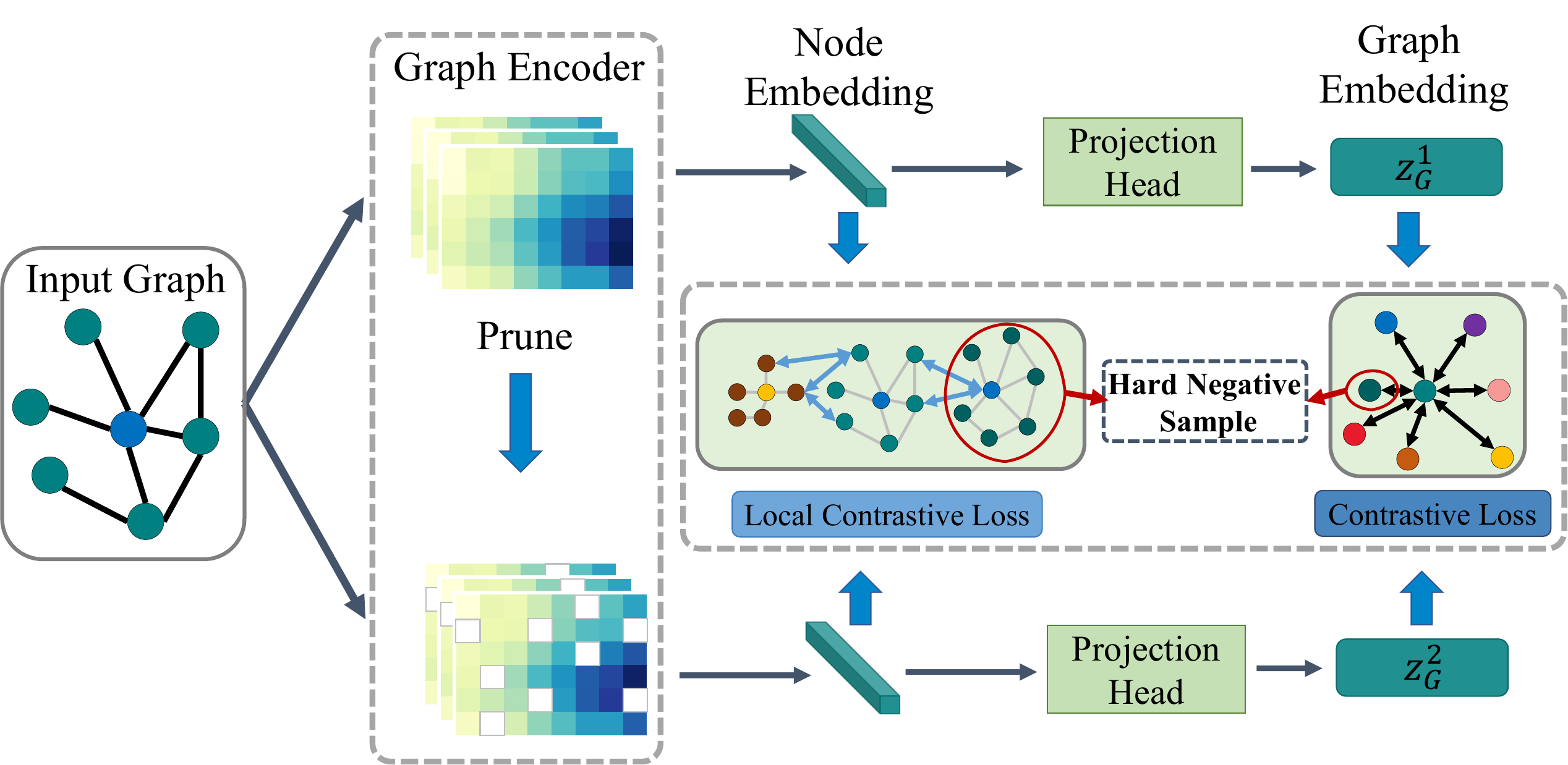}
\caption{\textbf{Framework of LAMP.} One branch takes the original graph as input instead of the augmented view. The other branch is pruned from the former online and also embeds the original graph. Besides the ubiquitous NT-Xent loss, the graph encoder is jointly optimized with a local contrastive loss to optimize the hypersphere of contrastive learning.}
\label{fig:framework}
\end{figure*}

Contrastive data augmentation on graphs presents a significant challenge compared to images due to the complex structural information and diverse contexts inherent in graph data~\cite{feng2022adversarial}.
Inspecting prior studies on graph contrastive learning, we can systematize the two common paradigms for view generation. The first category is the random or learnable data corruption, such as the four types of general augmentations (node dropping, edge perturbation, attribute masking, and subgraph) in GraphCL~\cite{you2020graph} and learnable edge dropping in AD-GCL~\cite{suresh2021adversarial}.
Despite the effectiveness of these graph views on various tasks, the proposed data augmentations via random corruption or learning suffer from structural damage and artificially introduced noise, which could alter the fundamental property of input graphs.
Based on predefined sub-structure substitution rules \cite{sun2021mocl} or contrasted with 3-dimensional geometric views \cite{liu2022pre,stark20223d}, the second way is to integrate the domain knowledge to alleviate the issue of semantic alteration with data corruption. However, the fusion of domain knowledge inevitably undermines the generalization of the pre-trained model to other domains \cite{li2022let}.

To fully leverage the potential of contrastive learning in the graph domain, it is desirable to develop a graph contrastive learning model that can preserve semantic information while remaining independent of domain-specific knowledge. 
To address this challenge, we shift our focus from contrastive views to the graph encoder within the contrastive learning framework. 
Drawing inspiration from model compression techniques~\cite{zhu2022adaptive}, we note that the performance of sparse sub-networks could be comparable to their complete versions~\cite{han2016deep_compression,chen2021unified}, suggesting that pruning may be a viable approach for graph contrastive learning. 
Accordingly, we introduce a novel framework called LAMP that enables Graph Contrastive \textbf{L}earning vi\textbf{A} \textbf{M}odel \textbf{P}runing to remedy these issues in previous works.

The framework of LAMP is shown in Figure~\ref{fig:framework}. LAMP takes the original graph as model input to prevent semantic information alteration from graph corruption. While fostering the model with the ability to identify the essential information, we employ pruning~\cite{han2016deep_compression,he2018soft} for perturbation.
In particular, since the pruned graph encoder is always obtained from the latest encoder, the two contrastive embeddings will co-evolve, which ensures model convergence.
Moreover, considering the hard negative samples that have different structural properties but output similar graph embedding, we develop a novel loss to enhance the contrastive learning with the node embeddings.
Despite the simplicity, coupling the two strategies together enable us to perform effective contrastive learning on graphs with model perturbation. 
Besides the superior performance of LAMP in the extensive experiments of unsupervised and transfer learning for graph classification, we also theoretically explain the superiority of model pruning compared to data augmentation.
The emerging contributions are elaborated below:

\begin{itemize}
\item We reformulate the framework of graph contrastive learning from model compression, which allows the model pre-training free from semantic information alteration and profound domain knowledge fusion.

\item Convinced by the capability of pruning in representation learning, we theoretically analyze the superiority of model pruning compared to data augmentation and present the instantiation, called LAMP.

\item Considering the correspondence of node embeddings in positive pair samples, we further enhance LAMP with the ability of handling hard negative samples by a local contrastive loss.

\item LAMP suppresses the SOTA competitors through extensive experiments in unsupervised and transfer learning.

\end{itemize}

\section{Related Works}
\label{sec:liter}
Graph contrastive learning has been broadly adopted for tasks without ground-truth labels, such as graph classification \cite{you2020graph,li2022let}, node classification~\cite{feng2022adversarial,hou2024ncd}, and link prediction~\cite{zhang2023line}. Here, we devote our attention to contrastive learning for graph classification, the most relevant topic of this work.

\subsection{Graph Contrastive Learning} 
Great success has been achieved by graph contrastive learning in self-supervised graph representation learning.
Among various research, the study of the contrastive view is a key issue in graph contrastive learning.
Recently, based on the data augmentation on images, numerous works have explored the feasibility of  augmentations on graphs~\cite{you2020graph,you2021graph,suresh2021adversarial,you2022bringing,li2022let}. 
While gratifying, the proposed data augmentations via random perturbation or learning suffer from structural damage and noisy information \cite{you2020graph,suresh2021adversarial}.
To tackle this issue, several works have attempted to preserve the graph semantic structure by resolving profound domain knowledge into augmentations, such as MoCL~\cite{sun2021mocl}, KCL~\cite{fang2022molecular}, GraphMVP~\cite{liu2022pre} and 3D-Infomax~\cite{stark20223d}
However, this domain knowledge only exists on molecules, which significantly limits the generality.
Moreover, beyond the general contrastive learning framework, DGCL~\cite{li2021disentangled} disentangles the graph encoder, and OEPG~\cite{yang2022omni} explores the semantic structure of datasets.
Although they present excellent performance, they still rely on the graph augmentations and thus are orthogonal to these works that keep the general contrastive learning framework; put differently, other models can work with the framework of DGCL and OEPG to produce superior performance. 
Among recent works, SimGRACE~\cite{xia2022simgrace} preserves semantics by disturbing the model weight with Gaussian noise. However, the introduced noise is data-agnostic, which could degenerate the model performance when the actual data distribution goes beyond the Gaussian distribution and explain the sub-optimal performance of SimGRACE in the experiment section.

\subsection{Model Pruning} 
In the early stage, pruning is generally a technique to improve model efficiency, which aims to shrink model size at surprisingly little sacrifice of model performance~\cite{lecun1989optimal}, and various pruning techniques have been proposed and effective for that goal~\cite{han2016deep_compression,li2017pruning,he2018soft}. 
Recently, besides the inherent function of model compression, several works explored its deeper connection with model generalization \cite{frankle2018lottery,zhang2021can}.
Moreover, the capability of pruning on model memorization has also been validated with long-tail distribution dataset~\cite{jiang2021self}.
In this paper, we particularly investigate contrastive learning via model pruning for graph representation learning. 
Note that, model pruning is first employed in this work to address the issue of semantics alteration caused by data augmentation.

\section{Notations and Preliminaries}
\label{sec:notations}
Before the elaboration of LAMP, here, some preliminary concepts and notations are given. Let $\mathcal{E}$ and $\mathcal{V}$ be the sets of edges and vertices, a simple graph can be formally written as $G=(\mathcal{V}, \mathcal{E})$.

\noindent \textbf{Graph representation learning.}
Generally, GNNs based on a message-passing scheme serve as the graph encoder~\cite{gilmer2017neural}. A GNN learns an embedding $h_v\in\mathbb{R}$ for each node and a vector $h_G\in\mathbb{R}$ is produced by a READOUT function for graph $G$. 
For an $L$-layer GNN, each node vector is decorated with the $L$-hop information from its neighbors. 
The hidden vector of node $v$ in layer $c$ is:
\begin{equation}
h_v^{(c)} = f_U^{(c)}(h_v^{(c\text{-}1)},f_M^{(c)}(\{(h_v^{(c\text{-}1)},h_u^{(c\text{-}1)})|u\in N(v)\})),
\label{eq:general_gnn}
\end{equation}
where $f_U^{(c)}$ aims to update each node vector in current layer, $f_M^{(c)}$ is the designed function for message-passing on graphs, the first-order neighbor nodes of $v$ is represented as $N(v)$, and $h_v^{(c)}$ denotes the hidden feature of $v$ in the $c$-th layer. After $L$ iterations, the entire graph representation can be written as
\begin{equation}
h_G = f_R(\{h_v|v\in\mathcal{V}\}),
\label{eq:gnn_readout}
\end{equation}
where $f_R$ pools the final set of node representations and is generally a summation or averaging function.

\noindent \textbf{Graph contrastive learning.}
A typical unsupervised model via contrastive learning takes two views from one graph as input, and the two views are produced by two data augmentation operators and serve as a positive pair.
At the phase for pre-training, a GNN-based encoder is used for structural information modeling of the input views, and a projection head is further employed to embed the two views into the same feature space for contrast.
Output feature vectors $h_i^1$ and $h_i^2$ from the same graph are expected to identify themselves from the others.
Thus, the NT-Xent loss \cite{chen2020simple} is adopted to achieve this goal via maximizing the consensus of a positive pair:
\begin{equation}
\mathcal{L}_i = -\log\frac{e^{sim(h_i^1, h_i^2)/\tau}}{\sum^N_{j=1,j\neq i}e^{sim(h_i^1, h_j^2)/\tau}},
\end{equation}
where $N$ is the batch size, $\tau$ controls the temperature parameter, and $sim(h^1, h^2)$ generally refers to a cosine similarity function $\frac{h^{1\top} h^2}{||h^1||\cdot||h^2||}$.
In particular, there are two types of negative pairs; put differently, $h_i^1$ can pair with all $h_j^2$, and $h_i^2$ can pair with all $h_j^1$. 

\noindent \textbf{The mutual information maximization principle.}
Graph contrastive learning leverages the principle of mutual information maximization (InfoMax) to enhance the correspondence between a graph representation and its corresponding views from various augmentations.
The graph representation $h_G$ is supposed to contain the feature underlying $G$, because the representation is expected to distinguish the graph $G$ from others within the same batch. The principle of mutual information maximization can formally be
\begin{equation}
\text{InfoMax:}\,\,\,\, \max I(G; h_G), where\,\, G\sim\mathbb{P}_\mathcal{G},
\end{equation}
where $\mathbb{P}_G$ denotes the distribution defined over the graph $G$ and $I(\cdot)$ refers to the mutual information.

\begin{figure}[!t]
  \begin{subfigure}{\linewidth}
    \centering
    \includegraphics[width=\linewidth]{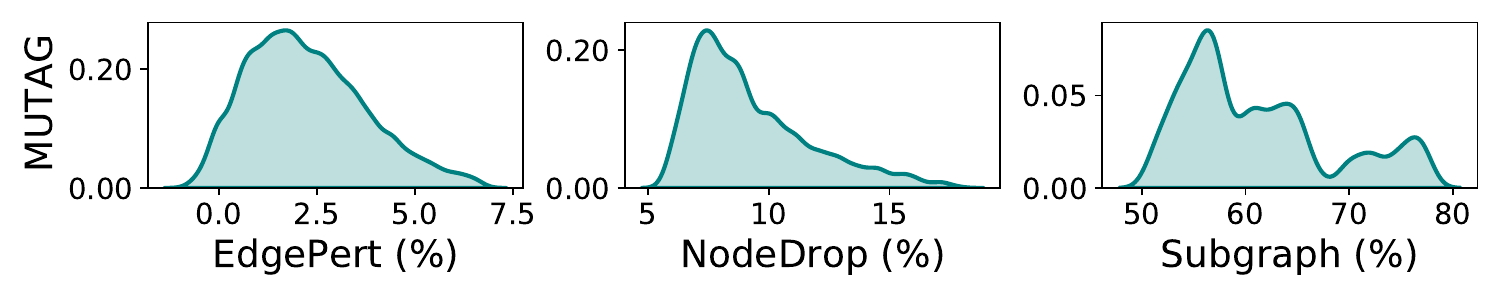}
  \end{subfigure}
  \\
  \begin{subfigure}{\linewidth}
    \centering
    \includegraphics[width=\linewidth]{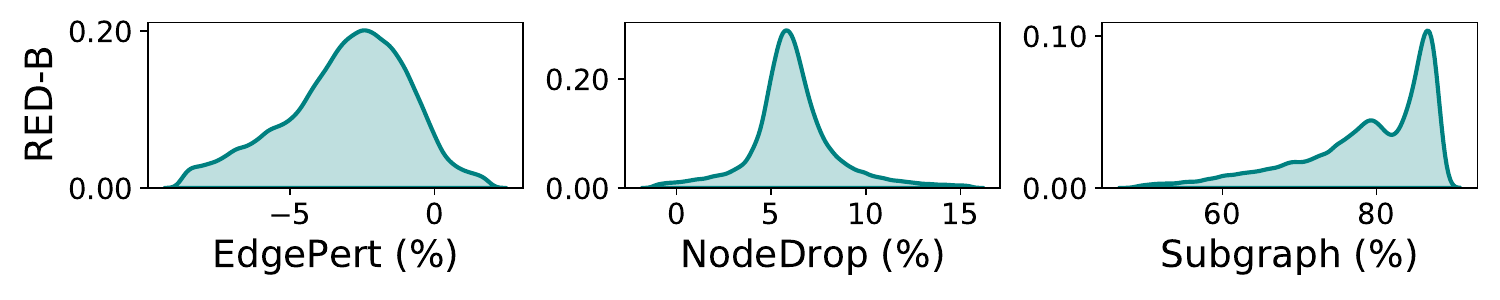}
  \end{subfigure}
  \caption{\textbf{Quantification of structural damage from data augmentation.} Percent change in structural entropy of MUTAG and REDDIT-BINARY after data augmentation (i.e., Edge perturbation, Node dropping, and Subgraph with 20\% strength from GraphCL).} 
  \label{fig:quantification_structural_damage}
\end{figure}

\section{Methodology}
\label{sec:methodology}
In this section, by turning the attention from data augmentation to model pruning, we bring about the proposed graph contrastive learning framework, termed \textit{LAMP}. Before the detailed description, we first give the motivation from the quantification of structural damage caused by general graph augmentation.

\subsection{Quantification of Structural Damage from Data Augmentation}
In previous works regarding graph contrastive learning, data augmentation is a general technique to help the graph encoder identify the essential information of input graphs~\cite{you2020graph,wu2022structural,yin2022autogcl}. Despite various forms, they are mostly built upon the concept of topology corruption, such as node dropping, edge perturbation, and learnable graph generation. Although these works have shown some extent of effectiveness on the actual tasks, the underlying structural damage remains, which inevitably leads to semantics alteration and undermines the model performance. 

Now, based on structural entropy~\cite{li2016structural}, we are the first to give the quantitative illustration of structural damage caused by data augmentations. 
Given a simple graph $G=(\mathcal{V}, \mathcal{E})$, the structural information underlying $G$ can be measured by:
\begin{equation}
\label{eq:structural_entropy}
\mathcal{H}(G)=-\sum_{v \in \mathcal{V}} \frac{g_{v}}{vol(\mathcal{V})} \log \frac{g_v}{vol(\mathcal{V})},
\end{equation}
in which $vol(\mathcal{V})$ denotes the sum of all node degrees and $g_v$ is the node degree of $v$. 
Among various forms of data augmentation proposed in previous works, here, we give the quantitative illustration of structural damage by three ubiquitous rules from GraphCL~\cite{you2020graph}, including subgraph, edge perturbation, and node dropping. In particular, the augmentation strength is consistent with the setting in GraphCL, that is, 20\%.
The structural damage of graph $G$ is measured by the percent change of its structural entropy. Formally, given any data augmentation function $t$, the percent change of structural entropy will be 
\begin{equation}
\mathcal{L}_{SE} = 1-\frac{\mathcal{H}(t(G))}{\mathcal{H}(G)}.
\end{equation}

The quantitative illustrations of structural damage caused by three data augmentation rules on a social network dataset (i.e., REDDIT-BINARY) and a bioinformatic dataset (i.e., MUTAG) are shown in Figure~\ref{fig:quantification_structural_damage}.~\footnote{The quantitative illustrations of other six datasets in unsupervised learning are shown in Appendix~\ref{sec:all_damage}.}
The effect of structural damage varies with the augmentation rules.
Specifically, node dropping and subgraph lead to different degrees of structural damage, and the information loss caused by subgraph is the largest and generally over 50\%. 
Besides the simple information loss, the structure damage composition of edge perturbation is more complex; put differently, in light of the distribution of REDDIT-BINARY, edge perturbation even introduces external data noise with the additional edges, which further interferes with the model from learning the actual structural information. Therefore, we naturally wonder: Can we design a more advanced model with effective contrastive pairs without structural damage? 
Next, to tackle our expectations, we present the superiority of model pruning in contrast to data augmentation.

\subsection{Theoretical Superiority of Model Pruning}
Through theoretical analysis, in this subsection, we present the property of graph encoder trained from model pruning.

\begin{theorem}
\label{theo:lamp-lowbound}
Suppose the graph encoder $f$ is implemented by a GNN with at least 2 layers and $f^*$ is the optimal version. Given a general data augmentation function $t$, the optimal pruned encoder $f_p^*$ satisfies,
\begin{itemize}
\item[1.] $I(f_p^*(G); Y) \ge I(f^*(t(G)); Y)$;
\item[2.] $I(f_p^*(G) ; f^*(G)) \geq I(f^*(t^1(G)) ; f^*(t^2(G)))$.
\end{itemize}
\end{theorem}

Statement 1 in Theorem~\ref{theo:lamp-lowbound} guarantees a lower bound of the mutual information between the learned representations and the labels of the downstream task; put differently, the learned essential information via the sparse encoder is more than views from augmentations.

Statement 2 in Theorem~\ref{theo:lamp-lowbound} suggests that the training performance of LAMP is better than the models based on augmentations in the architecture of graph contrastive learning.

\begin{proof}
Suppose $\mathbf{G}$ is a set of graphs. According to the definition, $f^\ast = argmax_{f} I(f(G);G)$, $f^\ast$ should be injective. Given some graph $G \in \mathbf{G}$, $G \Rightarrow f^\ast(G)$ is an injective deterministic mapping. Thus, for any random variable $Q$, 
\begin{equation}
I(f^\ast(G); Q) = I(G; Q). 
\end{equation}

When there is $Q = Y$, we will have,
\begin{equation}
I(f^\ast(G); Y) = I(G; Y).
\end{equation}

Moreover, in light of theoretically proof in \cite{malach2020proving}, a depth-two network can be approximated by pruning a random-weighted subnetwork $f^*_p$ as follows:
\begin{equation}
  \sup _{G \in \mathbf{G}}|f^\ast(G)-f^*_p(G)| \leq \epsilon.
\end{equation}

Accordingly, we have the following proof:
\begin{align}
  & I(f^*_p(G); Y ) - I(f^\ast(G); Y) \nonumber \\ 
= & H(f^*_p(G)) + H(Y) - H(f^*_p(G), Y) \nonumber \\ 
  & - H(f^\ast(G)) - H(Y) + H(f^\ast(G), Y) \nonumber \\
= & H(f^*_p(G))-H(f^\ast(G)) \nonumber \\
  & -(H(f^*_p(G), Y) - H(f^\ast(G), Y)) \nonumber \\
\stackrel{(a)}{\leq} & 2{\epsilon}^{2},
\end{align}
where $(a)$ is because of the arbitrariness of $\epsilon$ and the continuity of the entropy $H(\cdot)$. Meanwhile, because $\epsilon$ can be arbitrarily small, we can achieve
\begin{equation}
I(f^*_p(G); Y ) = I(f^\ast(G); Y). 
\end{equation}

Now, introducing the data processing inequality \cite{thomas2006elements} for data augmentation,
\begin{align}
I(f^\ast(G); Y) 
& = I(G; Y) \nonumber \\
& \geq I(t(G); Y) \nonumber \\
& = I(f^\ast(t(G)); Y). 
\end{align}

Combining above equations, we have the statement 1:
\begin{equation}
I(f^*_p(G); Y) = I(f^\ast (G); Y)  \geq I(f^\ast(t(G)); Y).
\label{eq:low-bound}
\end{equation}

Next, we come to proof the second statement,
\begin{align}
  & I(f^*(t^1(G)); f^*(t^2(G))) \nonumber\\
= & I(f^*(t^1(G)) ;(f^*(t^2(G)), Y)) \nonumber\\
& - I(f^*(t^1(G));(Y|f^*(t^2(G)))) \nonumber\\ 
\leq & I(f^*(t^1(G)) ;(f^*(t^2(G)), Y)) \nonumber\\
= & I(f^*(t^1(G));Y) \nonumber\\
  &+ I(f^*(t^1(G)) ;(f^*(t^2(G))|Y)).
\end{align}

Then, according to the data processing inequality \cite{thomas2006elements}, we move forward to
\begin{align}
& I(f^*(t^1(G));Y) + I(f^*(t^1(G)) ;(f^*(t^2(G))|Y)) \nonumber\\
\leq & I(f^*(G);Y) + I(f^*_p(t^1(G)) ;(f^*(t^2(G))|Y)) \nonumber\\
\leq & I(f^*(G);Y) + I(f^*_p(G) ;(f^*(G)|Y)) \nonumber\\
= & I(f^*_p(G);Y) + I(f^*_p(G) ;(f^*(G)|Y)) \nonumber\\
= & I(f^*_p(G) ;(f^*(G), Y)).
\end{align}

Finally, for the reason of $f^*_p(G) \perp_{f^*(G)} Y$,
\begin{align}
& I(f^*_p(G) ;(f^*(G), Y)) \nonumber\\
 = &I(f^*_p(G) ;(f^*(G), Y)) - I(f^*_p(G) ;(Y|f^*(G))) \nonumber\\
 = &I(f^*_p(G) ; f^*(G)),
\end{align}
which concludes the proof of the statement 2.
\end{proof}

\subsection{Instantiation of LAMP}
With the superiority of pruning in graph contrastive learning, now, we move from theory to practice.
Instead of producing contrastive pairs via data augmentation, here, we give our model design via model pruning. Figure~\ref{fig:framework} general pictures the workflow of LAMP. As can be seen, LAMP sticks to the twin-tower architecture of GraphCL~\cite{you2020graph}, while gets rid of its trail-and-error data augmentation rules. 
Especially, LAMP feeds the original graph into the dense graph encoder and its pruned version to contrast their embeddings.
Formally, let $G=(\mathbf{A}, \mathbf{X})$ be the input graph, $\mathbf{A}$ is the adjacency matrix, and $\mathbf{X}$ is the initial node features. 
For the $l$-th layer of graph encoder, the node representations of $G$ via Equation~\ref{eq:general_gnn} will be:
\begin{align}
H^{(l)}   &= f_U^{(l)}(f_M^{(l)}(\mathbf{A}, H^{(l-1)}); W^{(l)}), \\
H^{(l)}_p &= f_U^{(l)}(f_M^{(l)}(\mathbf{A}, H^{(l-1)}_p); p(W^{(l)})),
\end{align}
where $W^{(l)}$ is the weight matrix of update function $f_U^{(l)}$ and $p(\cdot)$ is the pruning function on model weight $W$.

In practice, we prune the dense graph encoder according to a pre-defined ratio $\gamma$ with the given pruning strategy, that is, the weight values of $W^{l}$ will be masked if they are ranked below $\gamma$.
In particular, the sparse degree of the graph encoder can be controlled by tuning the pruning ratio $\gamma$, which meets various demands of different datasets. Detailed discussion about the pruning ratio is conducted in the ablation study.
Moreover, to avoid drastic gradient changes and save computations, the wight matrix will be pruned at the beginning of each epoch. 
Since the sparse encoder is always obtained and updated from the latest dense version, the two branches will co-evolve during training.

\noindent \textbf{Remark.} To avoid possible confusion, we emphasize that the adoption of pruning is not aimed at enhancing model efficiency, but rather at boosting the performance of contrastive learning. Furthermore, our approach in this study is not tied to a specific pruning strategy, but is compatible with any given pruning methods. Specifically, we have implemented our method using two distinct pruning techniques: magnitude pruning~\cite{han2016deep_compression} and soft filter pruning~\cite{he2018soft}, which are referred to as \textbf{LAMP-Mag} and \textbf{LAMP-Soft}, respectively.

\noindent \textbf{Projection head.} After obtaining the graph representations via global pooling, a projection head $g(\cdot)$ is employed to cast the representations to another feature space for contrasting. In the framework of LAMP, a two-layer perceptron is also employed to produce the graph final representations $z^1_G$ and $z^2_G$,
\begin{equation}
z^1_G = g(h^1_G);\,\,\,\, z^2_G = g(h^2_G).
\end{equation}

\begin{figure}[!t]
  \centering
  \begin{subfigure}{0.45\linewidth}
    \centering
    \includegraphics[width=\linewidth]{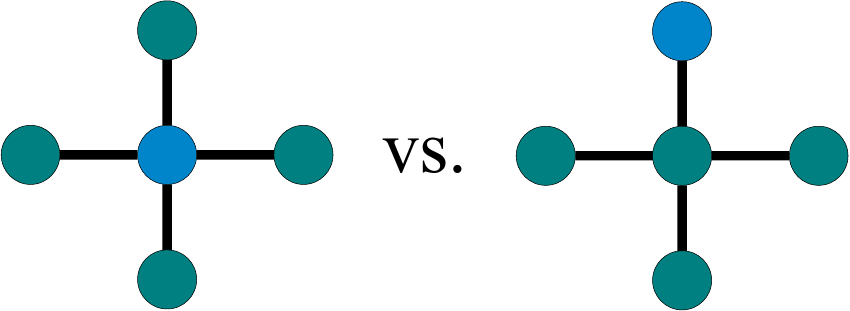}
    \caption{Same structure but different node features.}
    \label{fig:hard_negative_a}
  \end{subfigure}
  \begin{subfigure}{0.45\linewidth}
    \centering
    \includegraphics[width=\linewidth]{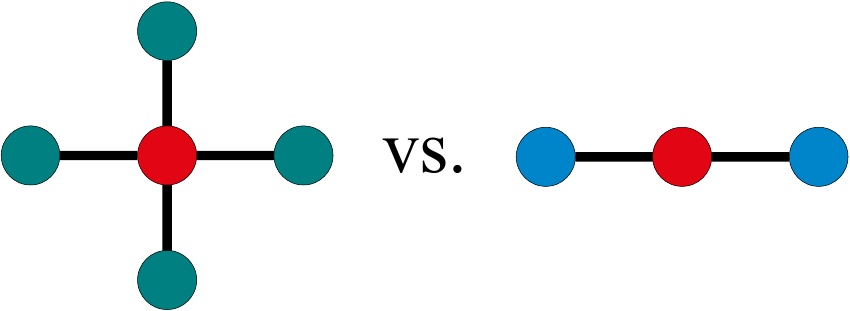}
    \caption{Different structure and different node features.}
    \label{fig:hard_negative_b}
  \end{subfigure}
  \caption{\textbf{Illustration of hard negative samples.} Via contrasting the graph embeddings, the pre-trained model is hard to distinguish this two kinds of graphs.} 
  \label{fig:hard_negative}
\end{figure}

\begin{table*}[!t]
\centering
\caption{Average accuracies (\%) $\pm$ Std. of LAMP and compared methods under the setting of unsupervised learning. \textbf{Bold} indicates the best performance over all methods. \underline{Underline} represents the second best. A.A. refers to the average accuracy over eight benchmarks. A.R. implies the abbreviation of average rank. The results of baselines are derived from the published works and - indicates the data missing in the such works.}
\label{tab:unsupervised}
\resizebox{\textwidth}{!}{%
\begin{tabular}{l|cccc|cccc|cc}
\hline \hline
 & NCI1 & PROTEINS & DD & MUTAG & COLLAB & RED-B & RED-M5K & IMDB-B & A.A. & A.R. \\ \hline \hline
GL & - & - & - & 81.66$\pm$2.11 & - & 77.34$\pm$0.18 & 41.01$\pm$0.17 & 65.87$\pm$0.98 & - & 15.5 \\
WL & 80.01$\pm$0.50 & 72.92$\pm$0.56 & - & 80.72$\pm$3.00 & - & 68.82$\pm$0.41 & 46.06$\pm$0.21 & 72.30$\pm$3.44 & - & 12.5 \\
DGK & 80.31$\pm$0.46 & 73.30$\pm$0.82 & - & 87.44$\pm$2.72 & - & 78.04$\pm$0.39 & 41.27$\pm$0.18 & 66.96$\pm$0.56 & - & 11.9 \\ \hline 
node2vec & 54.89$\pm$1.61 & 57.49$\pm$3.57 & - & 72.63$\pm$10.20 & - & - & - & - & - & 16.7 \\
sub2vec & 52.84$\pm$1.47 & 53.03$\pm$5.55 & - & 61.05$\pm$15.80 & - & 71.48$\pm$0.41 & 36.69$\pm$0.42 & 55.26$\pm$1.54 & - & 17.5 \\
graph2vec & 73.22$\pm$1.81 & 73.30$\pm$2.05 & - & 83.15$\pm$9.25 & - & 75.78$\pm$1.03 & 47.86$\pm$0.26 & 71.10$\pm$0.54 & - & 13.6 \\
MVGRL & - & - & - & 75.40$\pm$7.80 & - & 82.00$\pm$1.10 & - & 63.60$\pm$4.20 & - & 15.7 \\
InfoGraph & 76.20$\pm$1.06 & 74.44$\pm$0.31 & 72.85$\pm$1.78 & 89.01$\pm$1.13 & 70.65$\pm$1.13 & 82.50$\pm$1.42 & 53.46$\pm$1.03 & 73.03$\pm$0.87 & 74.02 & 9.8 \\
GraphCL & 77.87$\pm$0.41 & 74.39$\pm$0.45 & 78.62$\pm$0.40 & 86.80$\pm$1.34 & 71.36$\pm$1.15 & 89.53$\pm$0.84 & 55.99$\pm$0.28 & 71.14$\pm$0.44 & 75.71 & 8.9 \\
JOAO & 78.07$\pm$0.47 & 74.55$\pm$0.41 & 77.32$\pm$0.54 & 87.35$\pm$1.02 & 69.50$\pm$0.36 & 85.29$\pm$1.35 & 55.74$\pm$0.63 & 70.21$\pm$3.08 & 74.75 & 10.6 \\
JOAOv2 & 78.36$\pm$0.53 & 74.07$\pm$1.10 & 77.40$\pm$1.15 & 87.67$\pm$0.79 & 69.33$\pm$0.34 & 86.42$\pm$1.45 & 56.03$\pm$0.27 & 70.83$\pm$0.25 & 75.01 & 9.6 \\
AD-GCL & 73.91$\pm$0.77 & 73.28$\pm$0.47 & 75.79$\pm$0.87 & 88.74$\pm$1.85 & 72.02$\pm$0.56 & 90.07$\pm$0.85 & 54.33$\pm$0.32 & 70.21$\pm$0.68 & 74.79 & 10.2 \\
AutoGCL & 82.00$\pm$0.29 & 75.80$\pm$0.36 & 77.57$\pm$0.60 & 88.64$\pm$1.08 & 70.12$\pm$0.68 & 88.58$\pm$1.49 & 56.75$\pm$0.18 & 73.30$\pm$0.40 & 76.59 & 6.0 \\
RGCL & 78.14$\pm$1.08 & 75.03$\pm$0.43 & 78.86$\pm$0.48 & 87.66$\pm$1.01 & 70.92$\pm$0.65 & 90.34$\pm$0.58 & 56.38$\pm$0.40 & 71.85$\pm$0.84 & 76.15 & 6.6 \\
SimGRACE & 79.12$\pm$0.44 & 75.35$\pm$0.09 & 77.44$\pm$1.11 & 89.01$\pm$1.31 & 71.72$\pm$0.82 & 89.51$\pm$0.89 & 55.91$\pm$0.34 & 71.30$\pm$0.77 & 76.17 & 6.9 \\
SEGA & 79.00$\pm$0.72 & 76.01$\pm$0.42 & 78.76$\pm$0.57 & 90.21$\pm$0.66 & 74.12$\pm$0.47 & 90.21$\pm$0.65 & 56.13$\pm$0.30 & 73.58$\pm$0.44 & 77.25 & 4.3 \\
GCS & 77.37$\pm$0.30 & 75.02$\pm$0.39 & 77.22$\pm$0.30 & \underline{90.45$\pm$0.81} & \underline{75.56$\pm$0.41} & \textbf{92.98$\pm$0.28} & \underline{57.04$\pm$0.49} & 73.43$\pm$0.38 & 77.39 & 5.0 \\ \hline \hline
LAMP-Mag & \textbf{82.62$\pm$0.31} & \underline{76.75$\pm$0.67} & \underline{79.47$\pm$0.97} & 90.02$\pm$1.59 & 74.62$\pm$0.75 & 90.58$\pm$0.46 & 56.42$\pm$0.26 & \underline{73.46$\pm$0.65} & \underline{77.99} & \underline{2.8} \\ 
LAMP-Soft & \underline{82.17$\pm$0.48} & \textbf{77.34$\pm$0.53} & \textbf{80.03$\pm$0.85} & \textbf{90.89$\pm$1.04} & \textbf{75.96$\pm$0.67} & \underline{91.63$\pm$0.55} & \textbf{57.38$\pm$0.41} & \textbf{75.14$\pm$0.59} & \textbf{78.82} & \textbf{1.3} \\ \hline \hline
\end{tabular}%
}
\end{table*}

\subsection{Hard Negative Samples}
Among the research on contrastive learning, hard negative samples are quite ubiquitous and have great potential to improve model performance\cite{robinsoncontrastive}. However, little attention has been drawn to the hard negative samples within current contrastive learning for graph classification.
As shown in Figure~\ref{fig:hard_negative}, there are two kinds of hard negative samples.
In Figure~\ref{fig:hard_negative_a}, this negative pair has the same topology but different features. Let $h_{color}$ ($g$ for green, $b$ for blue) denote the node features, the graph representations would be similar after pooling as $f_R(4\times h_g + f_b) = f_R(4\times h_g + f_b)$.
In Figure~\ref{fig:hard_negative_b}, this negative pair has different topology and node features.
Let $h_b=2h_g$, the graph representations would be also similar through summation or maximization pooling function.
Given the common design of current contrastive learning for graph classification, the model is encouraged to enlarge the distance of negative pairs via graph representations, which may fail with the two kinds of negative samples in Figure~\ref{fig:hard_negative} and deteriorate model performance.

\begin{algorithm}[!t]
\caption{Pre-training algorithm of LAMP}
\label{code:lamp_training} 
\textbf{Input:} the training dataset $\mathbb{G} = \{G_1,G_2,\cdots\}$, graph encoder $f(\cdot)$ with weight matrix $\mathbf{W}_f$, projection head $g(\cdot)$ with weight matrix $\mathbf{W}_h$, pruning ratio $\gamma$ and learning rate $r$.

\begin{algorithmic}[1]
\STATE Initialize graph encoder $f(\cdot)$.
\FOR{each epoch} {
  \STATE Perform pruning to get $f_p(\cdot)$ with ratio $\gamma$;
  \FOR{each mini-batch}{
    \STATE Obtain the node representations $\mathbf{H}_\mathcal{V}^1$ and $\mathbf{H}_\mathcal{V}^2$ through the two graph encoders $f(\cdot)$ and $f_p(\cdot)$;
    \STATE Obtain the graph representations $\mathbf{Z}^1$ and $\mathbf{Z}^2$ through Equation~\ref{eq:gnn_readout} and projection head $g(\cdot)$;
    \STATE Calculate batch loss $\mathcal{L}$ based on Equation~\ref{eq:lamp_loss};
    \STATE Update weights: 
    \\ \qquad$\mathbf{W}_f' \leftarrow \mathbf{W}_f - r \triangledown_{\mathbf{W}_f}\mathcal{L}$
    \\ \qquad$\mathbf{W}_g' \leftarrow \mathbf{W}_g - r \triangledown_{\mathbf{W}_g}\mathcal{L}$
  }
  \ENDFOR
}
\ENDFOR
\end{algorithmic} 
\end{algorithm}

Despite the hardness of distinguishing the two kinds of negative samples from the graph representations, we may be able to spot some opportunities from the local features.
For example, the green nodes could be effortlessly identified from the blue nodes in Figure~\ref{fig:hard_negative_a}, and the read nodes also have obvious differences compared to the green and blue nodes in Figure~\ref{fig:hard_negative_b}.
Therefore, in order to cultivate the ability of contrastive learning to tackle hard negative samples, we propose a local contrastive loss with the node embeddings.

\noindent \textbf{Local Contrastive Loss.} Because each node embedding matrix after graph encoder contains the full set of nodes in original graph, here, we propose a local contrastive loss to enhance graph learning from the node level.
Our critical insight is that current models lack the ability to separate the hard negative samples via graph representations, while a contrastive angle based on node representations would be helpful in this scene.
Specifically, as the general NT-Xent loss enforces the dissimilarity among different graphs, we move forward to generalizing this dissimilarity to nodes. Formally, the local contrastive loss can be formulated as:
\begin{equation}
\mathcal{L}_{LocalC}^{v_i} = -\log\frac
{e^{sim(h_{v_i}^1, h_{v_i}^2)/\tau}}
{\sum_{\hat{G}\neq G}^{\mathbb{G}}\sum^{|\mathcal{V}_{\hat{G}}|}_{j=1}e^{sim(h_{v_i}^1, \hat{h}_{v_j}^2)/\tau}},
\end{equation}
where $\mathbb{G}$ refers to a training batch. Note that the local contrastive loss also has two kinds of negative pairs. In particular, since the node batch size varies among different datasets, LAMP will randomly sample a sub-node set of size $N_s$ for large dataset to avoid high computation costs. Considering the batch node size in actual model training, $N_s$ is fixed to 5,000 in the experiment setting.

\noindent \textbf{Remark.} Although the local contrastive loss for hard negative samples is relatively simple in its presentation, it is not easy to implement when the two graphs of a positive pair have different structures due to corruption. This limitation may account for why previous graph contrastive learning models have overlooked this approach and consequently produced sub-optimal performance.

Currently, we have presented the main components of the proposed LAMP that aims to help graph contrastive learning free from structural damage caused by data augmentation and enhance model training facing hard negative samples.
The final objective function of LAMP for pre-training is
\begin{equation}
\text{min}\,\,\mathcal{L} = \mathcal{L}_{G}+\alpha\mathcal{L}_{LocalC},
\label{eq:lamp_loss}
\end{equation}
where $\mathcal{L}_{G}$ is the NT-Xent loss, $\alpha$ controls the loss weight. Algorithm~\ref{code:lamp_training} reveals the pre-training procedure of LAMP.

\begin{table*}[!t]
\centering
\caption{Average test ROC-AUC (\%) $\pm$ Std. of LAMP along with baselines on eight downstream benchmarks under the setting of transfer learning. The results of baselines are derived from the corresponding works. \textbf{Bold} indicates the best performance among all baselines. \underline{Underline} gives the second best. Avg. shows the average ROC-AUC over all datasets. A.R. implies the abbreviation of average rank and - indicates the data missing in the such works.}
\label{tab:trans-results}
\resizebox{\textwidth}{!}{%
\begin{tabular}{l|cccccccc|cc}
\hline \hline
 & BBBP & Tox21 & ToxCast & SIDER & ClinTox & MUV & HIV & BACE & Avg. & A.R. \\ \hline \hline
No Pre-Train & 65.8$\pm$4.5 & 74.0$\pm$0.8 & 63.4$\pm$0.6 & 57.3$\pm$1.6 & 58.0$\pm$4.4 & 71.8$\pm$2.5 & 75.3$\pm$1.9 & 70.1$\pm$5.4 & 66.95 & 17.5 \\ \hline
Infomax & 68.8$\pm$0.8 & 75.3$\pm$0.6 & 62.7$\pm$0.4 & 58.4$\pm$0.8 & 69.9$\pm$3.0 & 75.3$\pm$2.5 & 76.0$\pm$0.7 & 75.9$\pm$1.6 & 70.29 & 15.3 \\
EdgePred & 67.3$\pm$2.4 & 76.0$\pm$0.6 & 64.1$\pm$0.6 & 60.4$\pm$0.7 & 64.1$\pm$3.7 & 74.1$\pm$2.1 & 76.3$\pm$1.0 & 79.9$\pm$0.9 & 70.28 & 12.7 \\
AttrMasking & 64.3$\pm$2.8 & 76.7$\pm$0.4 & 64.2$\pm$0.5 & 61.0$\pm$0.7 & 71.8$\pm$4.1 & 74.7$\pm$1.4 & 77.2$\pm$1.1 & 79.3$\pm$1.6 & 69.90 & 12.3 \\
ContextPred & 68.0$\pm$2.0 & 75.7$\pm$0.7 & 63.9$\pm$0.6 & 60.9$\pm$0.6 & 65.9$\pm$3.8 & 75.8$\pm$1.7 & 77.3$\pm$1.0 & 79.6$\pm$1.2 & 70.89 & 11.3 \\
GraphMAE & 72.0$\pm$0.6 & 75.5$\pm$0.6 & 64.1$\pm$0.3 & 60.3$\pm$1.1 & 82.3$\pm$1.2 & 76.3$\pm$2.4 & 77.2$\pm$1.0 & 83.1$\pm$0.9 & 73.85 & 8.5 \\
GraphMVP & 68.5$\pm$0.2 & 74.5$\pm$0.4 & 62.7$\pm$0.1 & 62.3$\pm$1.6 & 79.0$\pm$2.5 & 75.0$\pm$1.4 & 74.8$\pm$1.4 & 76.8$\pm$1.1 & 71.70 & 14.1 \\
GraphCL & 69.68$\pm$0.67 & 73.87$\pm$0.66 & 62.40$\pm$0.57 & 60.53$\pm$0.88 & 75.99$\pm$2.65 & 69.80$\pm$2.66 & 78.47$\pm$1.22 & 75.38$\pm$1.44 & 70.77 & 14.9 \\
JOAO & 70.22$\pm$0.98 & 74.98$\pm$0.29 & 62.94$\pm$0.48 & 59.97$\pm$0.79 & 81.32$\pm$2.49 & 71.66$\pm$1.43 & 76.73$\pm$1.23 & 77.34$\pm$0.48 & 71.89 & 13.9 \\
JOAOv2 & 71.39$\pm$0.92 & 74.27$\pm$0.62 & 63.16$\pm$0.45 & 60.49$\pm$0.74 & 80.97$\pm$1.64 & 73.67$\pm$1.00 & 77.51$\pm$1.17 & 75.49$\pm$1.27 & 72.12 & 12.9 \\
LP-Info & 71.40$\pm$0.55 & 74.54$\pm$0.45 & 63.04$\pm$0.30 & 59.70$\pm$0.43 & 74.81$\pm$2.73 & 72.99$\pm$2.28 & 76.96$\pm$1.10 & 80.21$\pm$1.36 & 71.71 & 13.4 \\
AD-GCL & 70.01$\pm$1.07 & 76.54$\pm$0.82 & 63.07$\pm$0.72 & 63.28$\pm$0.79 & 79.78$\pm$3.52 & 72.30$\pm$1.61 & 78.28$\pm$0.97 & 78.51$\pm$0.80 & 72.72 & 10.3 \\
AutoGCL & 73.36$\pm$0.77 & 75.69$\pm$0.29 & 63.47$\pm$0.38 & 62.51$\pm$0.63 & 80.99$\pm$3.38 & 75.83$\pm$1.30 & 78.35$\pm$0.64 & 83.26$\pm$1.13 & 74.18 & 6.9 \\
RGCL & 71.42$\pm$0.66 & 75.20$\pm$0.34 & 63.33$\pm$0.17 & 61.38$\pm$0.61 & 83.38$\pm$0.91 & 76.66$\pm$0.99 & 77.90$\pm$0.80 & 76.03$\pm$0.77 & 73.16 & 9.0 \\
D-SLA & 72.60$\pm$0.79 & 76.81$\pm$0.52 & 64.24$\pm$0.50 & 60.22$\pm$1.13 & 80.17$\pm$1.50 & 76.64$\pm$0.91 & 78.59$\pm$0.44 & 83.81$\pm$1.01 & 74.14 & 6.3 \\
SimGRACE & 71.25$\pm$0.86 & - & 63.36$\pm$0.52 & 60.59$\pm$0.96 & - & - & - & - & - & 11.3 \\
SEGA & 71.86$\pm$1.06 & 76.72$\pm$0.43 & 65.23$\pm$0.91 & 63.68$\pm$0.34 & 84.99$\pm$0.94 & 76.60$\pm$2.45 & 77.63$\pm$1.37 & 77.07$\pm$0.46 & 74.22 & 5.9 \\
GCS & 71.46$\pm$0.46 & 76.16$\pm$0.41 & \underline{65.35$\pm$0.17} & 64.20$\pm$0.35 & 82.01$\pm$1.90 & \textbf{80.45$\pm$1.67} & 80.22$\pm$1.37 & 77.90$\pm$0.26 & 74.72 & 4.9 \\ \hline \hline
LAMP-Mag & \underline{74.72$\pm$1.24} & \underline{76.86$\pm$0.31} & 64.92$\pm$0.55 & \textbf{64.85$\pm$0.73} & \underline{85.18$\pm$2.12} & 78.91$\pm$2.55 & \underline{80.38$\pm$0.75} & \underline{84.72$\pm$1.77} & \underline{76.32} & \underline{2.3} \\ 
LAMP-Soft & \textbf{75.77$\pm$0.76} & \textbf{77.23$\pm$0.41} & \textbf{65.87$\pm$0.33} & \underline{64.24$\pm$0.68} & \textbf{85.98$\pm$1.27} & \underline{79.50$\pm$2.19} & \textbf{81.73$\pm$1.25} & \textbf{85.58$\pm$1.43} & \textbf{76.99} & \textbf{1.3} \\ \hline \hline
\end{tabular}}
\end{table*}

\section{Experiments}
In this section, we are devoted to evaluating LAMP with extensive experiments. 
Under the setting of unsupervised and transfer learning, LAMP empirically shows its superiority compared to the SOTA competitors. 
Ablation studies regarding hyper-parameters are further conducted to make an in-depth analysis.

\subsection{Experimental Setup}
\textbf{Datasets.}
For unsupervised learning, various benchmarks are adopted from TUDataset~\cite{morris2020tudataset}, including COLLAB, REDDIT-BINARY, REDDIT-MULTI-5K, IMDB-BINARY, NCI1, MUTAG, PROTEINS and DD.
For transfer learning, ZINC15 \cite{sterling2015zinc} dataset is adopted for pre-training. In particular, a subset with two million unlabeled molecular graphs are sampled from the ZINC15.
We employ the eight ubiquitous benchmarks from the MoleculeNet dataset~\cite{wu2018moleculenet} as the downstream experiments regarding transfer learning. Further details are shown in Appendix~\ref{sec:data_summary}.

\noindent \textbf{Learning protocol.}
Following the learning setting in GraphCL~\cite{you2020graph}, the corresponding learning protocols are adopted for a fair comparison.
(a) In unsupervised representation learning, all data is used for model pre-training and the learned graph embeddings are then fed into a non-linear SVM classifier to perform 10-fold cross-validation.
(b) In transfer learning, we first pre-train the model on ZINC15. Then, we finetune and evaluate the model on MoleculeNet dataset using the scaffold split scheme~\cite{chen2012comparison}.

\noindent \textbf{Configuration.}
To keep in line with GraphCL~\cite{you2020graph}, the same GNN architectures are employed with their original hyper-parameters under individual experiment settings. 
Specifically, in unsupervised learning, GIN~\cite{xu2019powerful} with 32 hidden units and 3 layers is set up. 
In transfer learning, GIN is used with 5 layers and 300 hidden dimensions.
The pruning ratio for sparse encoder is selected from 5\% to 95\% with a step of 5\%. 
For local contrastive loss balance, $\alpha$ is tuned among $\{0.01, 0.1, 1, 10, 100\}$. 
Hyper-parameters are selected by the grid search on the validation sets.
Additional details are shown in the Appendix~\ref{sec:exp_setup}.

\noindent \textbf{Pruning strategy.} To demonstrate the compatibility of our framework, LAMP, with a variety of pruning methodologies, in this study, we adopt two distinct pruning techniques for the implementation of our method: magnitude pruning~\cite{han2016deep_compression} and soft filter pruning~\cite{he2018soft}, denoted as \textbf{LAMP-Mag} and \textbf{LAMP-Soft}, respectively.

\subsection{Results Compared with SOTAs}
\textbf{Unsupervised learning.}
The baselines in unsupervised learning have three categories. 
The first set is three SOTA kernel-based methods that include GL~\cite{shervashidze2009efficient}, WL~\cite{shervashidze2011weisfeiler}, and DGK~\cite{yanardag2015deep}. The second set is four heuristic self-supervised methods, including node2vec \cite{grover2016node2vec}, sub2vec \cite{adhikari2018sub2vec}, graph2vec \cite{narayanan2017graph2vec}, and InfoGraph \cite{sun2020infograph}. The final category comes from the graph contrastive learning domain, including MVGRL \cite{hassani2020contrastive}, GraphCL \cite{you2020graph}, AD-GCL~\cite{suresh2021adversarial}, JOAO~\cite{you2021graph}, AutoGCL~\cite{yin2022autogcl}, RGCL~\cite{li2022let}, SimGRACE~\cite{xia2022simgrace}, SEGA~\cite{wu2023sega} and GCS~\cite{wei2023boosting}.

The classification accuracies of LAMP against the SOTA competitors are shown in Table~\ref{tab:unsupervised}, and a significant performance improvement from the disappearance of the data augmentation can be witnessed as opposed to the baselines.
Before the specific performance description of LAMP, here, we first glance at SimGRACE, which first attempts to transfer the contrastive attention from data augmentation to model perturbation via introducing Gaussian noise to model weights. Although SimGRACE reveals its effectiveness on various datasets, the introduced Gaussian noise still degenerates the model performance; put differently, SimGRACE does not defeat the models with data augmentations.

We now present a comprehensive analysis of the superior performance of LAMP. As indicated by the final column for average rank, LAMP-Mag and LAMP-Soft secure the top two positions and exhibit the highest average accuracies among all baseline models. Notably, as evidenced by the column for average accuracy, LAMP-Soft surpasses the second-best model (i.e., GCS) with an accuracy improvement of 1.43\%. Specifically, LAMP-Soft achieves the best performance on six out of eight benchmarks, while still maintaining the second-best performance on the remaining two datasets. Although LAMP-Mag is not superior as LAMP-Soft, it achieves the best performance on the NCI1 dataset and the second-best on three other datasets.
In particular, we can observe a more significant improvement on bioinformatics datasets compared to social datasets. This discrepancy may be attributed to the presence of hard negative samples and our local contrastive loss may not perform optimally when the social datasets lack node features. 
In summary, we can conclude that model pruning may be a more promising direction for graph contrastive learning.

\begin{table}[!t]
\centering
\caption{The effectiveness of local contrastive loss $\mathcal{L}_{LocalC}$. A.A. is short for average accuracy.}
\label{tab:sensi_localc}
\resizebox{\linewidth}{!}{%
\begin{tabular}{l|ccc}
\hline \hline
 & SimGRACE & LAMP-Mag w/o $\mathcal{L}_{LocalC}$ & LAMP-Mag \\ \hline \hline
NCI1 & 79.12$\pm$0.44 & 80.26$\pm$0.58 & 82.62$\pm$0.31 \\
PROTEINS & 75.35$\pm$0.09 & 75.73$\pm$0.75 & 76.75$\pm$0.67 \\
DD & 77.44$\pm$1.11 & 78.44$\pm$0.79 & 79.47$\pm$0.97 \\
MUTAG & 89.01$\pm$1.31 & 89.88$\pm$1.51 & 90.02$\pm$1.59 \\
COLLAB & 71.72$\pm$0.82 & 72.65$\pm$0.51 & 74.62$\pm$0.75 \\
RED-B & 89.51$\pm$0.89 & 90.28$\pm$0.49 & 90.58$\pm$0.46 \\
RED-M5K & 55.91$\pm$0.34 & 55.69$\pm$0.36 & 56.42$\pm$0.26 \\
IMDB-B & 71.30$\pm$0.77 & 71.68$\pm$0.67 & 73.46$\pm$0.65 \\ \hline
A.A. & 76.17 & 76.83 & 77.99 \\ \hline \hline
\end{tabular}}
\end{table}

\noindent \textbf{Transfer learning.}
Baseline methods in transfer learning include EdgePred, AttrMsking, ContexPred \cite{hu2020strategies}, Infomax~\cite{velickovic2019deep}, JOAO~\cite{you2021graph}, GraphCL~\cite{you2020graph}, AD-GCL~\cite{suresh2021adversarial}, LP-Info~\cite{you2022bringing}, GraphMAE~\cite{hou2022graphmae}, AutoGCL~\cite{yin2022autogcl}, GraphMVP~\cite{liu2022pre}, RGCL~\cite{li2022let}, D-SLA~\cite{kim2022graph}, SimGRACE~\cite{xia2022simgrace}, SEGA~\cite{wu2023sega} and GCS~\cite{wei2023boosting}. A model without pre-train, termed `No Pre-Train', is also adopted for comparison.

The results of LAMP, along with baselines under the setting of transfer learning on eight benchmarks, are shown in Table~\ref{tab:trans-results}. 
In summary, the proposed models, namely LAMP-Mag and LAMP-Soft, demonstrate superior efficacy in comparison to preceding studies, as evidenced by the average ROC-AUC and ranking. Specifically, LAMP-Soft outperforms all other models on six out of the eight benchmarks, while securing second place on the remaining two datasets. LAMP-Mag exhibits optimal performance on the SIDER benchmark and obtains the second highest performance on five out of the eight benchmarks. In particular, LAMP-Soft obtains a 2.27\% performance gain in terms of average ROC-AUC compared to the best baselines (i.e., GCS with a 74.72\% average ROC-AUC) and over 10\% performance gain compared to the model trained from scratch.

\begin{figure}[!t]
  \centering
  \includegraphics[width=\linewidth]{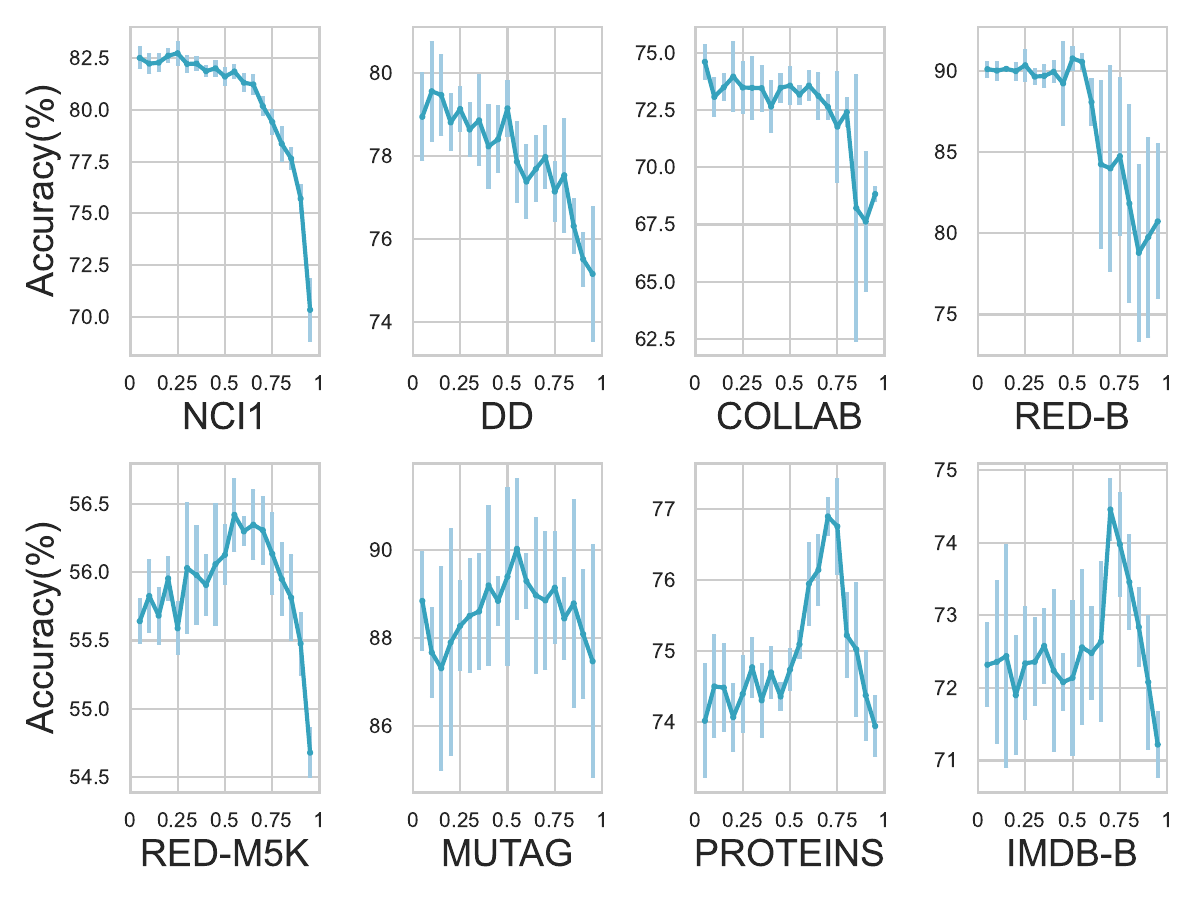}
  \caption{Sensitivity \textit{w.r.t.} pruning ratio $\gamma$.}
  \label{fig:sensi_prune_ratio}
\end{figure}

\subsection{Ablation Study}
\label{sec:ablation}
Here, we make an in-depth analysis about the performance of LAMP under the setting of unsupervised learning. In particular, the magnitude pruning is adopted for ablation study.

\noindent \textbf{Effectiveness of local contrastive loss.}
Besides the contrastive angle from the ubiquitous NT-Xent loss for unsupervised learning, we take another step to help model be capable of handling hard negative samples from the perspective of nodes. As shown in Table~\ref{tab:sensi_localc}, LAMP-Mag obtains better performance with a 1.16\% average accuracy gain when decorated with the proposed local contrastive loss, which suggests the effectiveness of the proposed local loss in addressing hard negative samples and improving model performance.
In particular, LAMP-Mag w/o $\mathcal{L}_{LocalC}$ has an average accuracy of 76.83\% on eight benchmarks, which not only outperforms current SOTAs for view generation but also defeats SimGRACE which disturbs the model weight with Gaussian noise. 
Thus, we can reaffirm the effectiveness of pruning on graph contrastive learning.

\noindent \textbf{Sensitivity regarding pruning ratio.}
The pruning ratio controls the information that the sparse encoder captures; thus, a proper ratio would help the model identify the essential structure of input graphs. As shown in Figure~\ref{fig:sensi_prune_ratio}, the datasets in the first row prefer a lower pruning ratio, while the other datasets would like a sparser encoder. Moreover, regardless of the optimal pruning ratio, the performances of all datasets quickly deteriorate when the sparsity goes above 75\%, due to limited capacity.

\begin{figure}[!t]
  \centering
  \includegraphics[width=\linewidth]{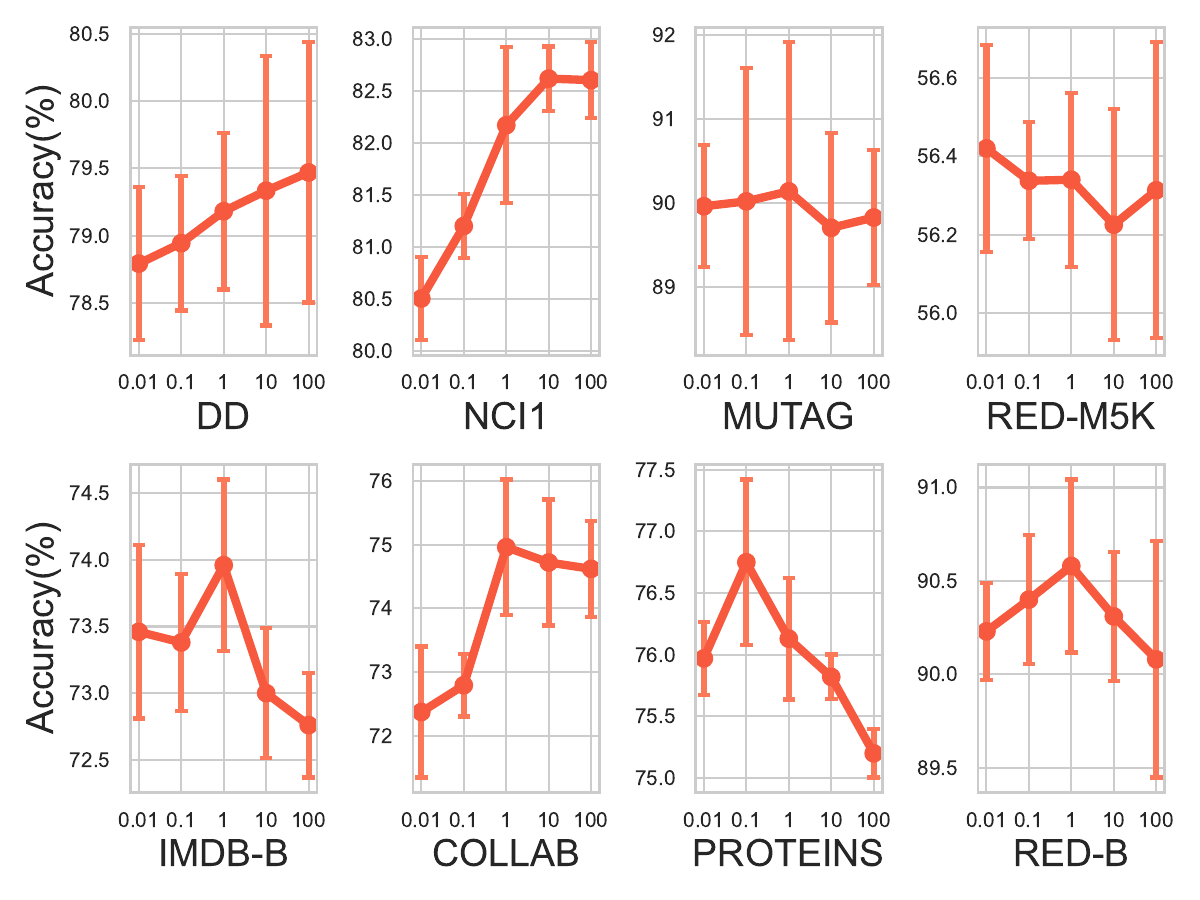}
  \caption{Sensitivity \textit{w.r.t.} loss balance $\alpha$.}
  \label{fig:sensi_alpha}
\end{figure}

\noindent \textbf{Sensitivity regarding loss balance.}
As we have validated the effectiveness of the proposed local contrastive loss, we further inspect the influence of its hyper-parameter (i.e., $\alpha$) on model performance. The unsupervised results of LAMP-Mag with candidate $\alpha$ on eight benchmarks are shown in Figure~\ref{fig:sensi_alpha}. As can be seen, DD and NCI1 show a positive correlation with $\alpha$, while MUTAG is not sensitive to the choice of $\alpha$, which is consistent with the stable performance of MUTAG w/o $\mathcal{L}_{LocalC}$ in Table~\ref{tab:sensi_localc}.
The other datasets show a trade-off within the given selections, and generally obtain the best performance with $\alpha$ around 1.

A further comprehensive analysis from Figure~\ref{fig:sensi_prune_ratio} and Figure~\ref{fig:sensi_alpha} shows that a suitable pruning ratio is particularly important for datasets such as PROTEIN, RED-M5K, and IMDB-B, whereas datasets like NCI1 and DD depend more on local contrastive loss.
The varying preferences indeed reflect the unique characteristics of different datasets. Social datasets may rely on pruning, potentially due to information redundancy. Conversely, datasets such as NCI1 and DD might depend more on local contrastive loss, which may be because the characteristics of bioinformatics datasets are more evident in local functional groups, and the efficacy of local contrastive loss on hard negative samples.

\section{Conclusion}
In this work, we reformulate the problem of graph contrastive learning from the angle of model compression.  
To avoid the loss of semantics caused by data augmentation, we present a novel method based on model pruning, termed \textit{LAMP}, rather than relying on profound domain knowledge.
Before the empirical validation, we theoretically explain the superiority of model pruning compared to data augmentation.
Extensive experiments under unsupervised and transfer learning show that LAMP suppresses the current SOTA methods based on data augmentations.
An automatic pruning ratio and more advanced pruning strategies shed light on the future research direction.
Furthermore, node classification and link prediction tasks are also important areas in graph-related tasks.
There are many excellent works in these two directions, and we see this as an opportunity for future exploration and learning.

\begin{acks}
This research was supported by NSFC (Grant No. 61932002).
\end{acks}

\bibliographystyle{ACM-Reference-Format}
\balance
\bibliography{myref}


\begin{thebibliography}{75}


\ifx \showCODEN    \undefined \def \showCODEN     #1{\unskip}     \fi
\ifx \showDOI      \undefined \def \showDOI       #1{#1}\fi
\ifx \showISBNx    \undefined \def \showISBNx     #1{\unskip}     \fi
\ifx \showISBNxiii \undefined \def \showISBNxiii  #1{\unskip}     \fi
\ifx \showISSN     \undefined \def \showISSN      #1{\unskip}     \fi
\ifx \showLCCN     \undefined \def \showLCCN      #1{\unskip}     \fi
\ifx \shownote     \undefined \def \shownote      #1{#1}          \fi
\ifx \showarticletitle \undefined \def \showarticletitle #1{#1}   \fi
\ifx \showURL      \undefined \def \showURL       {\relax}        \fi
\providecommand\bibfield[2]{#2}
\providecommand\bibinfo[2]{#2}
\providecommand\natexlab[1]{#1}
\providecommand\showeprint[2][]{arXiv:#2}

\bibitem[HIV({[n.\,d.]})]%
        {HIV}
 \bibinfo{year}{[n.\,d.]}\natexlab{}.
\newblock \bibinfo{title}{AIDS Antiviral Screen Data.}
  (\bibinfo{year}{[n.\,d.]}).
\newblock
\newblock
\shownote{Accessed: 2017-09-27,
  \url{https://wiki.nci.nih.gov/display/NCIDTPdata/AIDS+Antiviral+Screen+Data}}.


\bibitem[Tox(2014)]%
        {Tox21}
 \bibinfo{year}{2014}\natexlab{}.
\newblock \bibinfo{title}{Tox21 Data Challenge}.  (\bibinfo{year}{2014}).
\newblock
\newblock
\shownote{Accessed: 2017-09-27, \url{https://tripod.nih.gov/tox21/challenge}}.


\bibitem[Adhikari et~al\mbox{.}(2018)]%
        {adhikari2018sub2vec}
\bibfield{author}{\bibinfo{person}{Bijaya Adhikari}, \bibinfo{person}{Yao
  Zhang}, \bibinfo{person}{Naren Ramakrishnan}, {and} \bibinfo{person}{B~Aditya
  Prakash}.} \bibinfo{year}{2018}\natexlab{}.
\newblock \showarticletitle{Sub2vec: Feature learning for subgraphs}. In
  \bibinfo{booktitle}{\emph{Pacific-Asia Conference on Knowledge Discovery and
  Data Mining}}. Springer, \bibinfo{pages}{170--182}.
\newblock


\bibitem[Annamalai~Narayanan and Jaiswal(2017)]%
        {narayanan2017graph2vec}
\bibfield{author}{\bibinfo{person}{Rajasekar Venkatesan Lihui Chen Yang~Liu
  Annamalai~Narayanan, Mahinthan~Chandramohan} {and} \bibinfo{person}{Shantanu
  Jaiswal}.} \bibinfo{year}{2017}\natexlab{}.
\newblock \showarticletitle{graph2vec: Learning Distributed Representations of
  Graphs}. In \bibinfo{booktitle}{\emph{Proceedings of the 13th International
  Workshop on Mining and Learning with Graphs (MLG)}}.
\newblock


\bibitem[Bemis and Murcko(1996)]%
        {bemis1996properties}
\bibfield{author}{\bibinfo{person}{Guy~W Bemis} {and} \bibinfo{person}{Mark~A
  Murcko}.} \bibinfo{year}{1996}\natexlab{}.
\newblock \showarticletitle{The properties of known drugs. 1. Molecular
  frameworks}.
\newblock \bibinfo{journal}{\emph{Journal of Medicinal Chemistry}}
  \bibinfo{volume}{39}, \bibinfo{number}{15} (\bibinfo{year}{1996}),
  \bibinfo{pages}{2887--2893}.
\newblock


\bibitem[Chen et~al\mbox{.}(2012)]%
        {chen2012comparison}
\bibfield{author}{\bibinfo{person}{Bin Chen}, \bibinfo{person}{Robert~P
  Sheridan}, \bibinfo{person}{Viktor Hornak}, {and} \bibinfo{person}{Johannes~H
  Voigt}.} \bibinfo{year}{2012}\natexlab{}.
\newblock \showarticletitle{Comparison of random forest and Pipeline Pilot
  Naive Bayes in prospective QSAR predictions}.
\newblock \bibinfo{journal}{\emph{Journal of Chemical Information and
  Modeling}} \bibinfo{volume}{52}, \bibinfo{number}{3} (\bibinfo{year}{2012}),
  \bibinfo{pages}{792--803}.
\newblock


\bibitem[Chen et~al\mbox{.}(2020)]%
        {chen2020simple}
\bibfield{author}{\bibinfo{person}{Ting Chen}, \bibinfo{person}{Simon
  Kornblith}, \bibinfo{person}{Mohammad Norouzi}, {and}
  \bibinfo{person}{Geoffrey Hinton}.} \bibinfo{year}{2020}\natexlab{}.
\newblock \showarticletitle{A simple framework for contrastive learning of
  visual representations}. In \bibinfo{booktitle}{\emph{ICML}}. PMLR,
  \bibinfo{pages}{1597--1607}.
\newblock


\bibitem[Chen et~al\mbox{.}(2021)]%
        {chen2021unified}
\bibfield{author}{\bibinfo{person}{Tianlong Chen}, \bibinfo{person}{Yongduo
  Sui}, \bibinfo{person}{Xuxi Chen}, \bibinfo{person}{Aston Zhang}, {and}
  \bibinfo{person}{Zhangyang Wang}.} \bibinfo{year}{2021}\natexlab{}.
\newblock \showarticletitle{A unified lottery ticket hypothesis for graph
  neural networks}. In \bibinfo{booktitle}{\emph{International Conference on
  Machine Learning}}. PMLR, \bibinfo{pages}{1695--1706}.
\newblock


\bibitem[Fang et~al\mbox{.}(2022)]%
        {fang2022molecular}
\bibfield{author}{\bibinfo{person}{Yin Fang}, \bibinfo{person}{Qiang Zhang},
  \bibinfo{person}{Haihong Yang}, \bibinfo{person}{Xiang Zhuang},
  \bibinfo{person}{Shumin Deng}, \bibinfo{person}{Wen Zhang},
  \bibinfo{person}{Ming Qin}, \bibinfo{person}{Zhuo Chen},
  \bibinfo{person}{Xiaohui Fan}, {and} \bibinfo{person}{Huajun Chen}.}
  \bibinfo{year}{2022}\natexlab{}.
\newblock \showarticletitle{Molecular contrastive learning with chemical
  element knowledge graph}. In \bibinfo{booktitle}{\emph{Proceedings of the
  AAAI Conference on Artificial Intelligence}}, Vol.~\bibinfo{volume}{36}.
  \bibinfo{pages}{3968--3976}.
\newblock


\bibitem[Feng et~al\mbox{.}(2022)]%
        {feng2022adversarial}
\bibfield{author}{\bibinfo{person}{Shengyu Feng}, \bibinfo{person}{Baoyu Jing},
  \bibinfo{person}{Yada Zhu}, {and} \bibinfo{person}{Hanghang Tong}.}
  \bibinfo{year}{2022}\natexlab{}.
\newblock \showarticletitle{Adversarial graph contrastive learning with
  information regularization}. In \bibinfo{booktitle}{\emph{Proceedings of the
  ACM Web Conference 2022}}. \bibinfo{pages}{1362--1371}.
\newblock


\bibitem[Frankle and Carbin(2018)]%
        {frankle2018lottery}
\bibfield{author}{\bibinfo{person}{Jonathan Frankle} {and}
  \bibinfo{person}{Michael Carbin}.} \bibinfo{year}{2018}\natexlab{}.
\newblock \showarticletitle{The Lottery Ticket Hypothesis: Finding Sparse,
  Trainable Neural Networks}. In \bibinfo{booktitle}{\emph{International
  Conference on Learning Representations}}.
\newblock


\bibitem[Gardiner et~al\mbox{.}(2011)]%
        {gardiner2011effectiveness}
\bibfield{author}{\bibinfo{person}{Eleanor~J Gardiner}, \bibinfo{person}{John~D
  Holliday}, \bibinfo{person}{Caroline O’Dowd}, {and} \bibinfo{person}{Peter
  Willett}.} \bibinfo{year}{2011}\natexlab{}.
\newblock \showarticletitle{Effectiveness of 2D fingerprints for scaffold
  hopping}.
\newblock \bibinfo{journal}{\emph{Future Medicinal Chemistry}}
  \bibinfo{volume}{3}, \bibinfo{number}{4} (\bibinfo{year}{2011}),
  \bibinfo{pages}{405--414}.
\newblock


\bibitem[Gayvert et~al\mbox{.}(2016)]%
        {gayvert2016data}
\bibfield{author}{\bibinfo{person}{Kaitlyn~M Gayvert}, \bibinfo{person}{Neel~S
  Madhukar}, {and} \bibinfo{person}{Olivier Elemento}.}
  \bibinfo{year}{2016}\natexlab{}.
\newblock \showarticletitle{A data-driven approach to predicting successes and
  failures of clinical trials}.
\newblock \bibinfo{journal}{\emph{Cell Chemical Biology}} \bibinfo{volume}{23},
  \bibinfo{number}{10} (\bibinfo{year}{2016}), \bibinfo{pages}{1294--1301}.
\newblock


\bibitem[Gilmer et~al\mbox{.}(2017)]%
        {gilmer2017neural}
\bibfield{author}{\bibinfo{person}{Justin Gilmer}, \bibinfo{person}{Samuel~S
  Schoenholz}, \bibinfo{person}{Patrick~F Riley}, \bibinfo{person}{Oriol
  Vinyals}, {and} \bibinfo{person}{George~E Dahl}.}
  \bibinfo{year}{2017}\natexlab{}.
\newblock \showarticletitle{Neural message passing for quantum chemistry}. In
  \bibinfo{booktitle}{\emph{ICML}}. PMLR, \bibinfo{pages}{1263--1272}.
\newblock


\bibitem[Grover and Leskovec(2016)]%
        {grover2016node2vec}
\bibfield{author}{\bibinfo{person}{Aditya Grover} {and} \bibinfo{person}{Jure
  Leskovec}.} \bibinfo{year}{2016}\natexlab{}.
\newblock \showarticletitle{node2vec: Scalable feature learning for networks}.
\newblock \bibinfo{journal}{\emph{SIGKDD}} (\bibinfo{year}{2016}),
  \bibinfo{pages}{855--864}.
\newblock


\bibitem[Han et~al\mbox{.}(2016)]%
        {han2016deep_compression}
\bibfield{author}{\bibinfo{person}{Song Han}, \bibinfo{person}{Huizi Mao},
  {and} \bibinfo{person}{William~J Dally}.} \bibinfo{year}{2016}\natexlab{}.
\newblock \showarticletitle{Deep Compression: Compressing Deep Neural Networks
  with Pruning, Trained Quantization and Huffman Coding}.
\newblock \bibinfo{journal}{\emph{International Conference on Learning
  Representations (ICLR)}} (\bibinfo{year}{2016}).
\newblock


\bibitem[Hassani and Khasahmadi(2020)]%
        {hassani2020contrastive}
\bibfield{author}{\bibinfo{person}{Kaveh Hassani} {and}
  \bibinfo{person}{Amir~Hosein Khasahmadi}.} \bibinfo{year}{2020}\natexlab{}.
\newblock \showarticletitle{Contrastive multi-view representation learning on
  graphs}. In \bibinfo{booktitle}{\emph{ICML}}. PMLR,
  \bibinfo{pages}{4116--4126}.
\newblock


\bibitem[He et~al\mbox{.}(2020)]%
        {he2020momentum}
\bibfield{author}{\bibinfo{person}{Kaiming He}, \bibinfo{person}{Haoqi Fan},
  \bibinfo{person}{Yuxin Wu}, \bibinfo{person}{Saining Xie}, {and}
  \bibinfo{person}{Ross Girshick}.} \bibinfo{year}{2020}\natexlab{}.
\newblock \showarticletitle{Momentum contrast for unsupervised visual
  representation learning}. In \bibinfo{booktitle}{\emph{Proceedings of the
  IEEE/CVF Conference on Computer Vision and Pattern Recognition}}.
  \bibinfo{pages}{9729--9738}.
\newblock


\bibitem[He et~al\mbox{.}(2018)]%
        {he2018soft}
\bibfield{author}{\bibinfo{person}{Yang He}, \bibinfo{person}{Guoliang Kang},
  \bibinfo{person}{Xuanyi Dong}, \bibinfo{person}{Yanwei Fu}, {and}
  \bibinfo{person}{Yi Yang}.} \bibinfo{year}{2018}\natexlab{}.
\newblock \showarticletitle{Soft filter pruning for accelerating deep
  convolutional neural networks}. In \bibinfo{booktitle}{\emph{Proceedings of
  the 27th International Joint Conference on Artificial Intelligence}}.
  \bibinfo{pages}{2234--2240}.
\newblock


\bibitem[Hou et~al\mbox{.}(2024)]%
        {hou2024ncd}
\bibfield{author}{\bibinfo{person}{Yue Hou}, \bibinfo{person}{Xueyuan Chen},
  \bibinfo{person}{He Zhu}, \bibinfo{person}{Ruomei Liu},
  \bibinfo{person}{Bowen Shi}, \bibinfo{person}{Jiaheng Liu},
  \bibinfo{person}{Junran Wu}, {and} \bibinfo{person}{Ke Xu}.}
  \bibinfo{year}{2024}\natexlab{}.
\newblock \showarticletitle{NC$^{2}$D: Novel Class Discovery for Node
  Classification}. In \bibinfo{booktitle}{\emph{Proceedings of the 33st ACM
  International Conference on Information \& Knowledge Management}}.
\newblock


\bibitem[Hou et~al\mbox{.}(2022)]%
        {hou2022graphmae}
\bibfield{author}{\bibinfo{person}{Zhenyu Hou}, \bibinfo{person}{Xiao Liu},
  \bibinfo{person}{Yukuo Cen}, \bibinfo{person}{Yuxiao Dong},
  \bibinfo{person}{Hongxia Yang}, \bibinfo{person}{Chunjie Wang}, {and}
  \bibinfo{person}{Jie Tang}.} \bibinfo{year}{2022}\natexlab{}.
\newblock \showarticletitle{GraphMAE: Self-Supervised Masked Graph
  Autoencoders}. In \bibinfo{booktitle}{\emph{SIGKDD}}.
\newblock


\bibitem[Hu et~al\mbox{.}(2020)]%
        {hu2020strategies}
\bibfield{author}{\bibinfo{person}{W Hu}, \bibinfo{person}{B Liu},
  \bibinfo{person}{J Gomes}, \bibinfo{person}{M Zitnik}, \bibinfo{person}{P
  Liang}, \bibinfo{person}{V Pande}, {and} \bibinfo{person}{J Leskovec}.}
  \bibinfo{year}{2020}\natexlab{}.
\newblock \showarticletitle{Strategies For Pre-training Graph Neural Networks}.
\newblock \bibinfo{journal}{\emph{International Conference on Learning
  Representations (ICLR)}} (\bibinfo{year}{2020}).
\newblock


\bibitem[Jiang et~al\mbox{.}(2021)]%
        {jiang2021self}
\bibfield{author}{\bibinfo{person}{Ziyu Jiang}, \bibinfo{person}{Tianlong
  Chen}, \bibinfo{person}{Bobak~J Mortazavi}, {and} \bibinfo{person}{Zhangyang
  Wang}.} \bibinfo{year}{2021}\natexlab{}.
\newblock \showarticletitle{Self-damaging contrastive learning}. In
  \bibinfo{booktitle}{\emph{International Conference on Machine Learning}}.
  PMLR, \bibinfo{pages}{4927--4939}.
\newblock


\bibitem[Kim et~al\mbox{.}(2022)]%
        {kim2022graph}
\bibfield{author}{\bibinfo{person}{Dongki Kim}, \bibinfo{person}{Jinheon Baek},
  {and} \bibinfo{person}{Sung~Ju Hwang}.} \bibinfo{year}{2022}\natexlab{}.
\newblock \showarticletitle{Graph Self-supervised Learning with Accurate
  Discrepancy Learning}.
\newblock \bibinfo{journal}{\emph{Advances in Neural Information Processing
  Systems}} (\bibinfo{year}{2022}).
\newblock


\bibitem[Kingma and Ba(2015)]%
        {kingma2015adam}
\bibfield{author}{\bibinfo{person}{Diederik~P Kingma} {and}
  \bibinfo{person}{Jimmy Ba}.} \bibinfo{year}{2015}\natexlab{}.
\newblock \showarticletitle{Adam: A Method for Stochastic Optimization}. In
  \bibinfo{booktitle}{\emph{ICLR (Poster)}}.
\newblock


\bibitem[Kuhn et~al\mbox{.}(2016)]%
        {kuhn2016sider}
\bibfield{author}{\bibinfo{person}{Michael Kuhn}, \bibinfo{person}{Ivica
  Letunic}, \bibinfo{person}{Lars~Juhl Jensen}, {and} \bibinfo{person}{Peer
  Bork}.} \bibinfo{year}{2016}\natexlab{}.
\newblock \showarticletitle{The SIDER database of drugs and side effects}.
\newblock \bibinfo{journal}{\emph{Nucleic Acids Research}}
  \bibinfo{volume}{44}, \bibinfo{number}{D1} (\bibinfo{year}{2016}),
  \bibinfo{pages}{D1075--D1079}.
\newblock


\bibitem[Landrum(2013)]%
        {landrum2013rdkit}
\bibfield{author}{\bibinfo{person}{Greg Landrum}.}
  \bibinfo{year}{2013}\natexlab{}.
\newblock \showarticletitle{Rdkit documentation}.
\newblock \bibinfo{journal}{\emph{Release}} \bibinfo{volume}{1},
  \bibinfo{number}{1-79} (\bibinfo{year}{2013}).
\newblock


\bibitem[LeCun et~al\mbox{.}(1989)]%
        {lecun1989optimal}
\bibfield{author}{\bibinfo{person}{Yann LeCun}, \bibinfo{person}{John Denker},
  {and} \bibinfo{person}{Sara Solla}.} \bibinfo{year}{1989}\natexlab{}.
\newblock \showarticletitle{Optimal brain damage}.
\newblock \bibinfo{journal}{\emph{Advances in Neural Information Processing
  Systems}}  \bibinfo{volume}{2} (\bibinfo{year}{1989}).
\newblock


\bibitem[Li and Pan(2016)]%
        {li2016structural}
\bibfield{author}{\bibinfo{person}{Angsheng Li} {and} \bibinfo{person}{Yicheng
  Pan}.} \bibinfo{year}{2016}\natexlab{}.
\newblock \showarticletitle{Structural information and dynamical complexity of
  networks}.
\newblock \bibinfo{journal}{\emph{IEEE Transactions on Information Theory}}
  \bibinfo{volume}{62}, \bibinfo{number}{6} (\bibinfo{year}{2016}),
  \bibinfo{pages}{3290--3339}.
\newblock


\bibitem[Li et~al\mbox{.}(2017)]%
        {li2017pruning}
\bibfield{author}{\bibinfo{person}{Hao Li}, \bibinfo{person}{Asim Kadav},
  \bibinfo{person}{Igor Durdanovic}, \bibinfo{person}{Hanan Samet}, {and}
  \bibinfo{person}{Hans~Peter Graf}.} \bibinfo{year}{2017}\natexlab{}.
\newblock \showarticletitle{Pruning filters for efficient convnets}. In
  \bibinfo{booktitle}{\emph{International Conference on Machine Learning}}.
  PMLR.
\newblock


\bibitem[Li et~al\mbox{.}(2021)]%
        {li2021disentangled}
\bibfield{author}{\bibinfo{person}{Haoyang Li}, \bibinfo{person}{Xin Wang},
  \bibinfo{person}{Ziwei Zhang}, \bibinfo{person}{Zehuan Yuan},
  \bibinfo{person}{Hang Li}, {and} \bibinfo{person}{Wenwu Zhu}.}
  \bibinfo{year}{2021}\natexlab{}.
\newblock \showarticletitle{Disentangled contrastive learning on graphs}.
\newblock \bibinfo{journal}{\emph{Advances in Neural Information Processing
  Systems}}  \bibinfo{volume}{34} (\bibinfo{year}{2021}),
  \bibinfo{pages}{21872--21884}.
\newblock


\bibitem[Li et~al\mbox{.}(2022)]%
        {li2022let}
\bibfield{author}{\bibinfo{person}{Sihang Li}, \bibinfo{person}{Xiang Wang},
  \bibinfo{person}{An Zhang}, \bibinfo{person}{Yingxin Wu},
  \bibinfo{person}{Xiangnan He}, {and} \bibinfo{person}{Tat-Seng Chua}.}
  \bibinfo{year}{2022}\natexlab{}.
\newblock \showarticletitle{Let Invariant Rationale Discovery Inspire Graph
  Contrastive Learning}. In \bibinfo{booktitle}{\emph{International Conference
  on Machine Learning}}. PMLR, \bibinfo{pages}{13052--13065}.
\newblock


\bibitem[Liu et~al\mbox{.}(2022)]%
        {liu2022pre}
\bibfield{author}{\bibinfo{person}{Shengchao Liu}, \bibinfo{person}{Hanchen
  Wang}, \bibinfo{person}{Weiyang Liu}, \bibinfo{person}{Joan Lasenby},
  \bibinfo{person}{Hongyu Guo}, {and} \bibinfo{person}{Jian Tang}.}
  \bibinfo{year}{2022}\natexlab{}.
\newblock \showarticletitle{Pre-training Molecular Graph Representation with 3D
  Geometry}. In \bibinfo{booktitle}{\emph{International Conference on Learning
  Representations}}.
\newblock


\bibitem[Ma and Tang(2021)]%
        {ma2021deep}
\bibfield{author}{\bibinfo{person}{Yao Ma} {and} \bibinfo{person}{Jiliang
  Tang}.} \bibinfo{year}{2021}\natexlab{}.
\newblock \bibinfo{booktitle}{\emph{Deep learning on graphs}}.
\newblock \bibinfo{publisher}{Cambridge University Press}.
\newblock


\bibitem[Malach et~al\mbox{.}(2020)]%
        {malach2020proving}
\bibfield{author}{\bibinfo{person}{Eran Malach}, \bibinfo{person}{Gilad
  Yehudai}, \bibinfo{person}{Shai Shalev-Schwartz}, {and} \bibinfo{person}{Ohad
  Shamir}.} \bibinfo{year}{2020}\natexlab{}.
\newblock \showarticletitle{Proving the lottery ticket hypothesis: Pruning is
  all you need}. In \bibinfo{booktitle}{\emph{International Conference on
  Machine Learning}}. PMLR, \bibinfo{pages}{6682--6691}.
\newblock


\bibitem[Martins et~al\mbox{.}(2012)]%
        {martins2012bayesian}
\bibfield{author}{\bibinfo{person}{Ines~Filipa Martins}, \bibinfo{person}{Ana~L
  Teixeira}, \bibinfo{person}{Luis Pinheiro}, {and} \bibinfo{person}{Andre~O
  Falcao}.} \bibinfo{year}{2012}\natexlab{}.
\newblock \showarticletitle{A Bayesian approach to in silico blood-brain
  barrier penetration modeling}.
\newblock \bibinfo{journal}{\emph{Journal of Chemical Information and
  Modeling}} \bibinfo{volume}{52}, \bibinfo{number}{6} (\bibinfo{year}{2012}),
  \bibinfo{pages}{1686--1697}.
\newblock


\bibitem[Morris et~al\mbox{.}(2020)]%
        {morris2020tudataset}
\bibfield{author}{\bibinfo{person}{Christopher Morris},
  \bibinfo{person}{Nils~M. Kriege}, \bibinfo{person}{Franka Bause},
  \bibinfo{person}{Kristian Kersting}, \bibinfo{person}{Petra Mutzel}, {and}
  \bibinfo{person}{Marion Neumann}.} \bibinfo{year}{2020}\natexlab{}.
\newblock \showarticletitle{TUDataset: A collection of benchmark datasets for
  learning with graphs}.
\newblock \bibinfo{journal}{\emph{ICML 2020 Workshop on Graph Representation
  Learning and Beyond}} (\bibinfo{year}{2020}).
\newblock
\showeprint{2007.08663}


\bibitem[Novick et~al\mbox{.}(2013)]%
        {novick2013sweetlead}
\bibfield{author}{\bibinfo{person}{Paul~A Novick}, \bibinfo{person}{Oscar~F
  Ortiz}, \bibinfo{person}{Jared Poelman}, \bibinfo{person}{Amir~Y Abdulhay},
  {and} \bibinfo{person}{Vijay~S Pande}.} \bibinfo{year}{2013}\natexlab{}.
\newblock \showarticletitle{SWEETLEAD: an in silico database of approved drugs,
  regulated chemicals, and herbal isolates for computer-aided drug discovery}.
\newblock \bibinfo{journal}{\emph{PloS One}} \bibinfo{volume}{8},
  \bibinfo{number}{11} (\bibinfo{year}{2013}), \bibinfo{pages}{e79568}.
\newblock


\bibitem[Ramsundar et~al\mbox{.}(2019)]%
        {ramsundar2019deep}
\bibfield{author}{\bibinfo{person}{Bharath Ramsundar}, \bibinfo{person}{Peter
  Eastman}, \bibinfo{person}{Patrick Walters}, {and} \bibinfo{person}{Vijay
  Pande}.} \bibinfo{year}{2019}\natexlab{}.
\newblock \bibinfo{booktitle}{\emph{Deep learning for the life sciences:
  applying deep learning to genomics, microscopy, drug discovery, and more}}.
\newblock \bibinfo{publisher}{O'Reilly Media}.
\newblock


\bibitem[Richard et~al\mbox{.}(2016)]%
        {richard2016toxcast}
\bibfield{author}{\bibinfo{person}{Ann~M Richard}, \bibinfo{person}{Richard~S
  Judson}, \bibinfo{person}{Keith~A Houck}, \bibinfo{person}{Christopher~M
  Grulke}, \bibinfo{person}{Patra Volarath}, \bibinfo{person}{Inthirany
  Thillainadarajah}, \bibinfo{person}{Chihae Yang}, \bibinfo{person}{James
  Rathman}, \bibinfo{person}{Matthew~T Martin}, \bibinfo{person}{John~F
  Wambaugh}, {et~al\mbox{.}}} \bibinfo{year}{2016}\natexlab{}.
\newblock \showarticletitle{ToxCast chemical landscape: paving the road to 21st
  century toxicology}.
\newblock \bibinfo{journal}{\emph{Chemical Research in Toxicology}}
  \bibinfo{volume}{29}, \bibinfo{number}{8} (\bibinfo{year}{2016}),
  \bibinfo{pages}{1225--1251}.
\newblock


\bibitem[Robinson et~al\mbox{.}(2021)]%
        {robinsoncontrastive}
\bibfield{author}{\bibinfo{person}{Joshua~David Robinson},
  \bibinfo{person}{Ching-Yao Chuang}, \bibinfo{person}{Suvrit Sra}, {and}
  \bibinfo{person}{Stefanie Jegelka}.} \bibinfo{year}{2021}\natexlab{}.
\newblock \showarticletitle{Contrastive Learning with Hard Negative Samples}.
  In \bibinfo{booktitle}{\emph{International Conference on Learning
  Representations}}.
\newblock


\bibitem[Sheridan(2013)]%
        {sheridan2013time}
\bibfield{author}{\bibinfo{person}{Robert~P Sheridan}.}
  \bibinfo{year}{2013}\natexlab{}.
\newblock \showarticletitle{Time-split cross-validation as a method for
  estimating the goodness of prospective prediction.}
\newblock \bibinfo{journal}{\emph{Journal of Chemical Information and
  Modeling}} (\bibinfo{year}{2013}).
\newblock


\bibitem[Shervashidze et~al\mbox{.}(2011)]%
        {shervashidze2011weisfeiler}
\bibfield{author}{\bibinfo{person}{Nino Shervashidze}, \bibinfo{person}{Pascal
  Schweitzer}, \bibinfo{person}{Erik~Jan Van~Leeuwen}, \bibinfo{person}{Kurt
  Mehlhorn}, {and} \bibinfo{person}{Karsten~M Borgwardt}.}
  \bibinfo{year}{2011}\natexlab{}.
\newblock \showarticletitle{Weisfeiler-lehman graph kernels.}
\newblock \bibinfo{journal}{\emph{Journal of Machine Learning Research}}
  \bibinfo{volume}{12}, \bibinfo{number}{9} (\bibinfo{year}{2011}).
\newblock


\bibitem[Shervashidze et~al\mbox{.}(2009)]%
        {shervashidze2009efficient}
\bibfield{author}{\bibinfo{person}{Nino Shervashidze}, \bibinfo{person}{SVN
  Vishwanathan}, \bibinfo{person}{Tobias Petri}, \bibinfo{person}{Kurt
  Mehlhorn}, {and} \bibinfo{person}{Karsten Borgwardt}.}
  \bibinfo{year}{2009}\natexlab{}.
\newblock \showarticletitle{Efficient graphlet kernels for large graph
  comparison}. In \bibinfo{booktitle}{\emph{Artificial Intelligence and
  Statistics}}. PMLR, \bibinfo{pages}{488--495}.
\newblock


\bibitem[St{\"a}rk et~al\mbox{.}(2022)]%
        {stark20223d}
\bibfield{author}{\bibinfo{person}{Hannes St{\"a}rk},
  \bibinfo{person}{Dominique Beaini}, \bibinfo{person}{Gabriele Corso},
  \bibinfo{person}{Prudencio Tossou}, \bibinfo{person}{Christian Dallago},
  \bibinfo{person}{Stephan G{\"u}nnemann}, {and} \bibinfo{person}{Pietro
  Li{\`o}}.} \bibinfo{year}{2022}\natexlab{}.
\newblock \showarticletitle{3d infomax improves gnns for molecular property
  prediction}. In \bibinfo{booktitle}{\emph{International Conference on Machine
  Learning}}. PMLR, \bibinfo{pages}{20479--20502}.
\newblock


\bibitem[Sterling and Irwin(2015)]%
        {sterling2015zinc}
\bibfield{author}{\bibinfo{person}{Teague Sterling} {and}
  \bibinfo{person}{John~J Irwin}.} \bibinfo{year}{2015}\natexlab{}.
\newblock \showarticletitle{ZINC 15--ligand discovery for everyone}.
\newblock \bibinfo{journal}{\emph{Journal of Chemical Information and
  Modeling}} \bibinfo{volume}{55}, \bibinfo{number}{11} (\bibinfo{year}{2015}),
  \bibinfo{pages}{2324--2337}.
\newblock


\bibitem[Subramanian et~al\mbox{.}(2016)]%
        {subramanian2016computational}
\bibfield{author}{\bibinfo{person}{Govindan Subramanian},
  \bibinfo{person}{Bharath Ramsundar}, \bibinfo{person}{Vijay Pande}, {and}
  \bibinfo{person}{Rajiah~Aldrin Denny}.} \bibinfo{year}{2016}\natexlab{}.
\newblock \showarticletitle{Computational modeling of $\beta$-secretase 1
  (BACE-1) inhibitors using ligand based approaches}.
\newblock \bibinfo{journal}{\emph{Journal of Chemical Information and
  Modeling}} \bibinfo{volume}{56}, \bibinfo{number}{10} (\bibinfo{year}{2016}),
  \bibinfo{pages}{1936--1949}.
\newblock


\bibitem[Sun et~al\mbox{.}(2020)]%
        {sun2020infograph}
\bibfield{author}{\bibinfo{person}{Fan-Yun Sun}, \bibinfo{person}{Jordon
  Hoffman}, \bibinfo{person}{Vikas Verma}, {and} \bibinfo{person}{Jian Tang}.}
  \bibinfo{year}{2020}\natexlab{}.
\newblock \showarticletitle{InfoGraph: Unsupervised and Semi-supervised
  Graph-Level Representation Learning via Mutual Information Maximization}.
\newblock \bibinfo{journal}{\emph{ICLR}} (\bibinfo{year}{2020}).
\newblock


\bibitem[Sun et~al\mbox{.}(2021)]%
        {sun2021mocl}
\bibfield{author}{\bibinfo{person}{Mengying Sun}, \bibinfo{person}{Jing Xing},
  \bibinfo{person}{Huijun Wang}, \bibinfo{person}{Bin Chen}, {and}
  \bibinfo{person}{Jiayu Zhou}.} \bibinfo{year}{2021}\natexlab{}.
\newblock \showarticletitle{MoCL: data-driven molecular fingerprint via
  knowledge-aware contrastive learning from molecular graph}. In
  \bibinfo{booktitle}{\emph{Proceedings of the 27th ACM SIGKDD Conference on
  Knowledge Discovery \& Data Mining}}. \bibinfo{pages}{3585--3594}.
\newblock


\bibitem[Suresh et~al\mbox{.}(2021)]%
        {suresh2021adversarial}
\bibfield{author}{\bibinfo{person}{Susheel Suresh}, \bibinfo{person}{Pan Li},
  \bibinfo{person}{Cong Hao}, {and} \bibinfo{person}{Jennifer Neville}.}
  \bibinfo{year}{2021}\natexlab{}.
\newblock \showarticletitle{Adversarial graph augmentation to improve graph
  contrastive learning}.
\newblock \bibinfo{journal}{\emph{Advances in Neural Information Processing
  Systems}}  \bibinfo{volume}{34} (\bibinfo{year}{2021}).
\newblock


\bibitem[Thomas and Joy(2006)]%
        {thomas2006elements}
\bibfield{author}{\bibinfo{person}{MTCAJ Thomas} {and}
  \bibinfo{person}{A~Thomas Joy}.} \bibinfo{year}{2006}\natexlab{}.
\newblock \bibinfo{booktitle}{\emph{Elements of information theory}}.
\newblock \bibinfo{publisher}{Wiley-Interscience}.
\newblock


\bibitem[Tian et~al\mbox{.}(2020)]%
        {tian2020makes}
\bibfield{author}{\bibinfo{person}{Yonglong Tian}, \bibinfo{person}{Chen Sun},
  \bibinfo{person}{Ben Poole}, \bibinfo{person}{Dilip Krishnan},
  \bibinfo{person}{Cordelia Schmid}, {and} \bibinfo{person}{Phillip Isola}.}
  \bibinfo{year}{2020}\natexlab{}.
\newblock \showarticletitle{What makes for good views for contrastive
  learning?}
\newblock \bibinfo{journal}{\emph{Advances in Neural Information Processing
  Systems}} (\bibinfo{year}{2020}).
\newblock


\bibitem[Velickovic et~al\mbox{.}(2019)]%
        {velickovic2019deep}
\bibfield{author}{\bibinfo{person}{Petar Velickovic}, \bibinfo{person}{William
  Fedus}, \bibinfo{person}{William~L Hamilton}, \bibinfo{person}{Pietro
  Li{\`o}}, \bibinfo{person}{Yoshua Bengio}, {and} \bibinfo{person}{R~Devon
  Hjelm}.} \bibinfo{year}{2019}\natexlab{}.
\newblock \showarticletitle{Deep Graph Infomax.}
\newblock \bibinfo{journal}{\emph{ICLR (Poster)}} \bibinfo{volume}{2},
  \bibinfo{number}{3} (\bibinfo{year}{2019}), \bibinfo{pages}{4}.
\newblock


\bibitem[Wei et~al\mbox{.}(2023)]%
        {wei2023boosting}
\bibfield{author}{\bibinfo{person}{Chunyu Wei}, \bibinfo{person}{Yu Wang},
  \bibinfo{person}{Bing Bai}, \bibinfo{person}{Kai Ni}, \bibinfo{person}{David
  Brady}, {and} \bibinfo{person}{Lu Fang}.} \bibinfo{year}{2023}\natexlab{}.
\newblock \showarticletitle{Boosting graph contrastive learning via graph
  contrastive saliency}. In \bibinfo{booktitle}{\emph{International conference
  on machine learning}}. PMLR, \bibinfo{pages}{36839--36855}.
\newblock


\bibitem[Wu et~al\mbox{.}(2023)]%
        {wu2023sega}
\bibfield{author}{\bibinfo{person}{Junran Wu}, \bibinfo{person}{Xueyuan Chen},
  \bibinfo{person}{Bowen Shi}, \bibinfo{person}{Shangzhe Li}, {and}
  \bibinfo{person}{Ke Xu}.} \bibinfo{year}{2023}\natexlab{}.
\newblock \showarticletitle{Sega: Structural entropy guided anchor view for
  graph contrastive learning}. In \bibinfo{booktitle}{\emph{International
  Conference on Machine Learning}}. PMLR, \bibinfo{pages}{37293--37312}.
\newblock


\bibitem[Wu et~al\mbox{.}(2022a)]%
        {wu2022structural}
\bibfield{author}{\bibinfo{person}{Junran Wu}, \bibinfo{person}{Xueyuan Chen},
  \bibinfo{person}{Ke Xu}, {and} \bibinfo{person}{Shangzhe Li}.}
  \bibinfo{year}{2022}\natexlab{a}.
\newblock \showarticletitle{Structural entropy guided graph hierarchical
  pooling}. In \bibinfo{booktitle}{\emph{International conference on machine
  learning}}. PMLR, \bibinfo{pages}{24017--24030}.
\newblock


\bibitem[Wu et~al\mbox{.}(2022b)]%
        {wu2022simple}
\bibfield{author}{\bibinfo{person}{Junran Wu}, \bibinfo{person}{Shangzhe Li},
  \bibinfo{person}{Jianhao Li}, \bibinfo{person}{Yicheng Pan}, {and}
  \bibinfo{person}{Ke Xu}.} \bibinfo{year}{2022}\natexlab{b}.
\newblock \showarticletitle{A simple yet effective method for graph
  classification}. In \bibinfo{booktitle}{\emph{Proceedings of the Thirty-First
  International Joint Conference on Artificial Intelligence, {IJCAI} 2022,
  Vienna, Austria, July 23-29, 2022}}. \bibinfo{publisher}{ijcai.org}.
\newblock


\bibitem[Wu et~al\mbox{.}(2018a)]%
        {wu2018moleculenet}
\bibfield{author}{\bibinfo{person}{Zhenqin Wu}, \bibinfo{person}{Bharath
  Ramsundar}, \bibinfo{person}{Evan~N Feinberg}, \bibinfo{person}{Joseph
  Gomes}, \bibinfo{person}{Caleb Geniesse}, \bibinfo{person}{Aneesh~S Pappu},
  \bibinfo{person}{Karl Leswing}, {and} \bibinfo{person}{Vijay Pande}.}
  \bibinfo{year}{2018}\natexlab{a}.
\newblock \showarticletitle{MoleculeNet: a benchmark for molecular machine
  learning}.
\newblock \bibinfo{journal}{\emph{Chemical Science}} \bibinfo{volume}{9},
  \bibinfo{number}{2} (\bibinfo{year}{2018}), \bibinfo{pages}{513--530}.
\newblock


\bibitem[Wu et~al\mbox{.}(2018b)]%
        {wu2018unsupervised}
\bibfield{author}{\bibinfo{person}{Zhirong Wu}, \bibinfo{person}{Yuanjun
  Xiong}, \bibinfo{person}{Stella~X Yu}, {and} \bibinfo{person}{Dahua Lin}.}
  \bibinfo{year}{2018}\natexlab{b}.
\newblock \showarticletitle{Unsupervised feature learning via non-parametric
  instance discrimination}. In \bibinfo{booktitle}{\emph{Proceedings of the
  IEEE Conference on Computer Vision and Pattern Recognition}}.
  \bibinfo{pages}{3733--3742}.
\newblock


\bibitem[Xia et~al\mbox{.}(2022)]%
        {xia2022simgrace}
\bibfield{author}{\bibinfo{person}{Jun Xia}, \bibinfo{person}{Lirong Wu},
  \bibinfo{person}{Jintao Chen}, \bibinfo{person}{Bozhen Hu}, {and}
  \bibinfo{person}{Stan~Z Li}.} \bibinfo{year}{2022}\natexlab{}.
\newblock \showarticletitle{SimGRACE: A Simple Framework for Graph Contrastive
  Learning without Data Augmentation}. In \bibinfo{booktitle}{\emph{Proceedings
  of the ACM Web Conference 2022}}. \bibinfo{pages}{1070--1079}.
\newblock


\bibitem[Xu et~al\mbox{.}(2019)]%
        {xu2019powerful}
\bibfield{author}{\bibinfo{person}{Keyulu Xu}, \bibinfo{person}{Weihua Hu},
  \bibinfo{person}{Jure Leskovec}, {and} \bibinfo{person}{Stefanie Jegelka}.}
  \bibinfo{year}{2019}\natexlab{}.
\newblock \showarticletitle{How Powerful are Graph Neural Networks?}. In
  \bibinfo{booktitle}{\emph{ICLR}}.
\newblock


\bibitem[Yanardag and Vishwanathan(2015)]%
        {yanardag2015deep}
\bibfield{author}{\bibinfo{person}{Pinar Yanardag} {and} \bibinfo{person}{SVN
  Vishwanathan}.} \bibinfo{year}{2015}\natexlab{}.
\newblock \showarticletitle{Deep graph kernels}.
\newblock \bibinfo{journal}{\emph{SIGKDD}} (\bibinfo{year}{2015}),
  \bibinfo{pages}{1365--1374}.
\newblock


\bibitem[Yang and Hong(2022)]%
        {yang2022omni}
\bibfield{author}{\bibinfo{person}{Ling Yang} {and} \bibinfo{person}{Shenda
  Hong}.} \bibinfo{year}{2022}\natexlab{}.
\newblock \showarticletitle{Omni-Granular Ego-Semantic Propagation for
  Self-Supervised Graph Representation Learning}. In
  \bibinfo{booktitle}{\emph{International Conference on Machine Learning}}.
  PMLR, \bibinfo{pages}{25022--25037}.
\newblock


\bibitem[Yang et~al\mbox{.}(2019)]%
        {yang2019xlnet}
\bibfield{author}{\bibinfo{person}{Zhilin Yang}, \bibinfo{person}{Zihang Dai},
  \bibinfo{person}{Yiming Yang}, \bibinfo{person}{Jaime Carbonell},
  \bibinfo{person}{Russ~R Salakhutdinov}, {and} \bibinfo{person}{Quoc~V Le}.}
  \bibinfo{year}{2019}\natexlab{}.
\newblock \showarticletitle{Xlnet: Generalized autoregressive pretraining for
  language understanding}.
\newblock \bibinfo{journal}{\emph{Advances in neural information processing
  systems}}  \bibinfo{volume}{32} (\bibinfo{year}{2019}).
\newblock


\bibitem[Yin et~al\mbox{.}(2022)]%
        {yin2022autogcl}
\bibfield{author}{\bibinfo{person}{Yihang Yin}, \bibinfo{person}{Qingzhong
  Wang}, \bibinfo{person}{Siyu Huang}, \bibinfo{person}{Haoyi Xiong}, {and}
  \bibinfo{person}{Xiang Zhang}.} \bibinfo{year}{2022}\natexlab{}.
\newblock \showarticletitle{AutoGCL: Automated graph contrastive learning via
  learnable view generators}. In \bibinfo{booktitle}{\emph{Proceedings of the
  AAAI Conference on Artificial Intelligence}}, Vol.~\bibinfo{volume}{36}.
  \bibinfo{pages}{8892--8900}.
\newblock


\bibitem[You et~al\mbox{.}(2021)]%
        {you2021graph}
\bibfield{author}{\bibinfo{person}{Yuning You}, \bibinfo{person}{Tianlong
  Chen}, \bibinfo{person}{Yang Shen}, {and} \bibinfo{person}{Zhangyang Wang}.}
  \bibinfo{year}{2021}\natexlab{}.
\newblock \showarticletitle{Graph contrastive learning automated}. In
  \bibinfo{booktitle}{\emph{ICML}}. PMLR, \bibinfo{pages}{12121--12132}.
\newblock


\bibitem[You et~al\mbox{.}(2020)]%
        {you2020graph}
\bibfield{author}{\bibinfo{person}{Yuning You}, \bibinfo{person}{Tianlong
  Chen}, \bibinfo{person}{Yongduo Sui}, \bibinfo{person}{Ting Chen},
  \bibinfo{person}{Zhangyang Wang}, {and} \bibinfo{person}{Yang Shen}.}
  \bibinfo{year}{2020}\natexlab{}.
\newblock \showarticletitle{Graph contrastive learning with augmentations}.
\newblock \bibinfo{journal}{\emph{Advances in Neural Information Processing
  Systems}}  \bibinfo{volume}{33} (\bibinfo{year}{2020}),
  \bibinfo{pages}{5812--5823}.
\newblock


\bibitem[You et~al\mbox{.}(2022)]%
        {you2022bringing}
\bibfield{author}{\bibinfo{person}{Yuning You}, \bibinfo{person}{Tianlong
  Chen}, \bibinfo{person}{Zhangyang Wang}, {and} \bibinfo{person}{Yang Shen}.}
  \bibinfo{year}{2022}\natexlab{}.
\newblock \showarticletitle{Bringing Your Own View: Graph Contrastive Learning
  without Prefabricated Data Augmentations} \emph{(\bibinfo{series}{WSDM
  '22})}. \bibinfo{publisher}{Association for Computing Machinery},
  \bibinfo{address}{New York, NY, USA}, \bibinfo{pages}{1300–1309}.
\newblock


\bibitem[Zhang et~al\mbox{.}(2021)]%
        {zhang2021can}
\bibfield{author}{\bibinfo{person}{Dinghuai Zhang}, \bibinfo{person}{Kartik
  Ahuja}, \bibinfo{person}{Yilun Xu}, \bibinfo{person}{Yisen Wang}, {and}
  \bibinfo{person}{Aaron Courville}.} \bibinfo{year}{2021}\natexlab{}.
\newblock \showarticletitle{Can subnetwork structure be the key to
  out-of-distribution generalization?}. In
  \bibinfo{booktitle}{\emph{International Conference on Machine Learning}}.
  PMLR, \bibinfo{pages}{12356--12367}.
\newblock


\bibitem[Zhang et~al\mbox{.}(2023)]%
        {zhang2023line}
\bibfield{author}{\bibinfo{person}{Zehua Zhang}, \bibinfo{person}{Shilin Sun},
  \bibinfo{person}{Guixiang Ma}, {and} \bibinfo{person}{Caiming Zhong}.}
  \bibinfo{year}{2023}\natexlab{}.
\newblock \showarticletitle{Line Graph Contrastive Learning for Link
  Prediction}.
\newblock \bibinfo{journal}{\emph{Pattern Recognition}} (\bibinfo{year}{2023}),
  \bibinfo{pages}{109537}.
\newblock


\bibitem[Zhu et~al\mbox{.}(2024b)]%
        {zhu2024hill}
\bibfield{author}{\bibinfo{person}{He Zhu}, \bibinfo{person}{Junran Wu},
  \bibinfo{person}{Ruomei Liu}, \bibinfo{person}{Yue Hou}, \bibinfo{person}{Ze
  Yuan}, \bibinfo{person}{Shangzhe Li}, \bibinfo{person}{Yicheng Pan}, {and}
  \bibinfo{person}{Ke Xu}.} \bibinfo{year}{2024}\natexlab{b}.
\newblock \showarticletitle{HILL: Hierarchy-aware Information Lossless
  Contrastive Learning for Hierarchical Text Classification}. In
  \bibinfo{booktitle}{\emph{Proceedings of the 2024 Conference of the North
  American Chapter of the Association for Computational Linguistics: Human
  Language Technologies (Volume 1: Long Papers)}}. \bibinfo{pages}{4731--4745}.
\newblock


\bibitem[Zhu et~al\mbox{.}(2023b)]%
        {zhu2023hitin}
\bibfield{author}{\bibinfo{person}{He Zhu}, \bibinfo{person}{Chong Zhang},
  \bibinfo{person}{Junjie Huang}, \bibinfo{person}{Junran Wu}, {and}
  \bibinfo{person}{Ke Xu}.} \bibinfo{year}{2023}\natexlab{b}.
\newblock \showarticletitle{HiTIN: Hierarchy-aware Tree Isomorphism Network for
  Hierarchical Text Classification}. In \bibinfo{booktitle}{\emph{Proceedings
  of the 61st Annual Meeting of the Association for Computational Linguistics
  (Volume 1: Long Papers)}}. \bibinfo{pages}{7809--7821}.
\newblock


\bibitem[Zhu et~al\mbox{.}(2024a)]%
        {zhu2024do}
\bibfield{author}{\bibinfo{person}{Jiaqi Zhu}, \bibinfo{person}{Shaofeng Cai},
  \bibinfo{person}{Fang Deng}, {and} \bibinfo{person}{WuJunran}.}
  \bibinfo{year}{2024}\natexlab{a}.
\newblock \showarticletitle{Do {LLM}s Understand Visual Anomalies? Uncovering
  {LLM}'s Capabilities in Zero-shot Anomaly Detection}. In
  \bibinfo{booktitle}{\emph{ACM Multimedia 2024}}.
\newblock
\urldef\tempurl%
\url{https://openreview.net/forum?id=JyOGUqYrbV}
\showURL{%
\tempurl}


\bibitem[Zhu et~al\mbox{.}(2022)]%
        {zhu2022adaptive}
\bibfield{author}{\bibinfo{person}{Jiaqi Zhu}, \bibinfo{person}{Fang Deng},
  \bibinfo{person}{Jiachen Zhao}, {and} \bibinfo{person}{Jie Chen}.}
  \bibinfo{year}{2022}\natexlab{}.
\newblock \showarticletitle{Adaptive aggregation-distillation autoencoder for
  unsupervised anomaly detection}.
\newblock \bibinfo{journal}{\emph{Pattern Recognition}}  \bibinfo{volume}{131}
  (\bibinfo{year}{2022}), \bibinfo{pages}{108897}.
\newblock


\bibitem[Zhu et~al\mbox{.}(2023a)]%
        {zhu2023uaed}
\bibfield{author}{\bibinfo{person}{Jiaqi Zhu}, \bibinfo{person}{Fang Deng},
  \bibinfo{person}{Jiachen Zhao}, \bibinfo{person}{Daoming Liu}, {and}
  \bibinfo{person}{Jie Chen}.} \bibinfo{year}{2023}\natexlab{a}.
\newblock \showarticletitle{Uaed: Unsupervised abnormal emotion detection
  network based on wearable mobile device}.
\newblock \bibinfo{journal}{\emph{IEEE Transactions on Network Science and
  Engineering}} \bibinfo{volume}{10}, \bibinfo{number}{6}
  (\bibinfo{year}{2023}), \bibinfo{pages}{3682--3696}.
\newblock


\end{thebibliography}

\clearpage
\appendix

\setcounter{table}{0}
\setcounter{figure}{0}
\renewcommand{\thetable}{A.\arabic{table}}
\renewcommand{\thefigure}{A.\arabic{figure}}

\section{Quantification of Structural Damage from Data Augmentation}
\label{sec:all_damage}
\begin{figure}[!hp]
  \centering\begin{subfigure}{\linewidth}
    \centering
    \includegraphics[width=\linewidth]{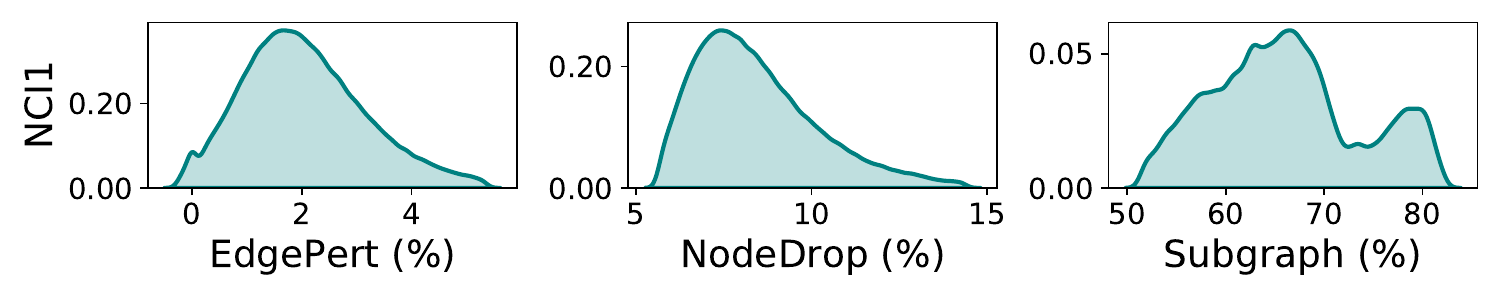}
  \end{subfigure}
  \\
  \begin{subfigure}{\linewidth}
    \centering
    \includegraphics[width=\linewidth]{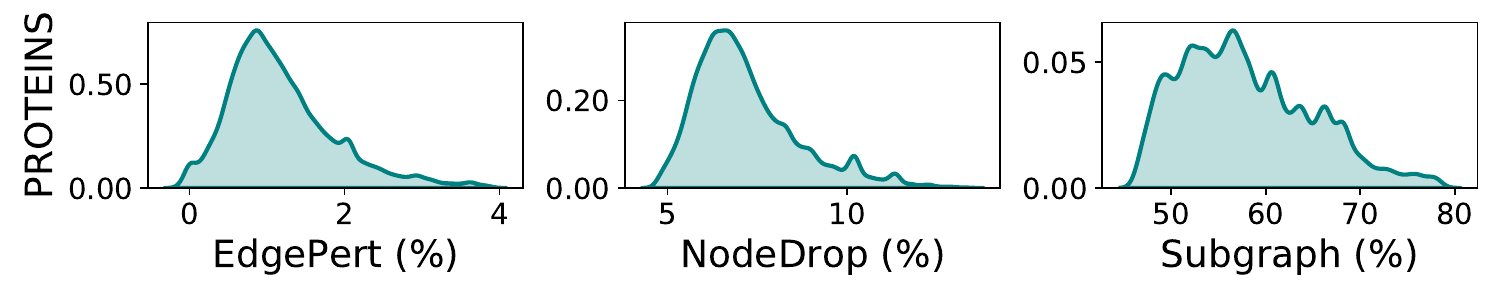}
  \end{subfigure}
  \\
  \begin{subfigure}{\linewidth}
    \centering
    \includegraphics[width=\linewidth]{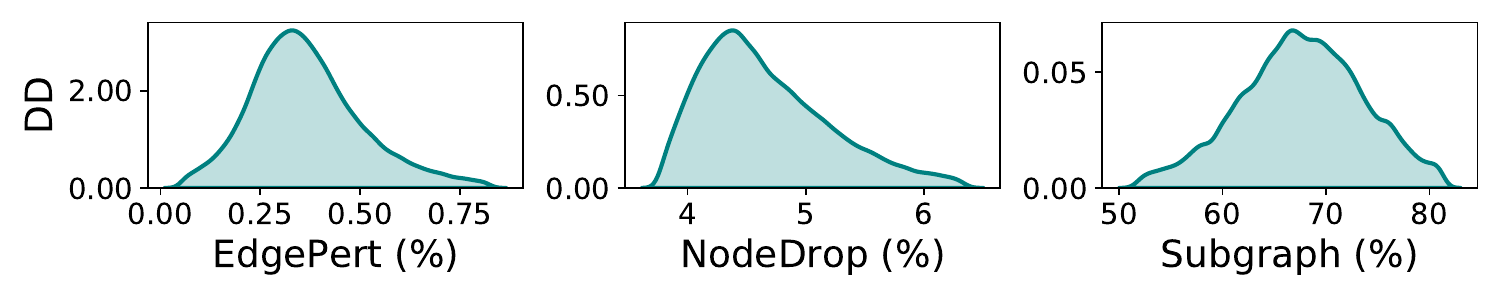}
  \end{subfigure}
  \\
  \begin{subfigure}{\linewidth}
    \centering
    \includegraphics[width=\linewidth]{MUTAG_aug.pdf}
  \end{subfigure}
  \\
  \begin{subfigure}{\linewidth}
    \centering
    \includegraphics[width=\linewidth]{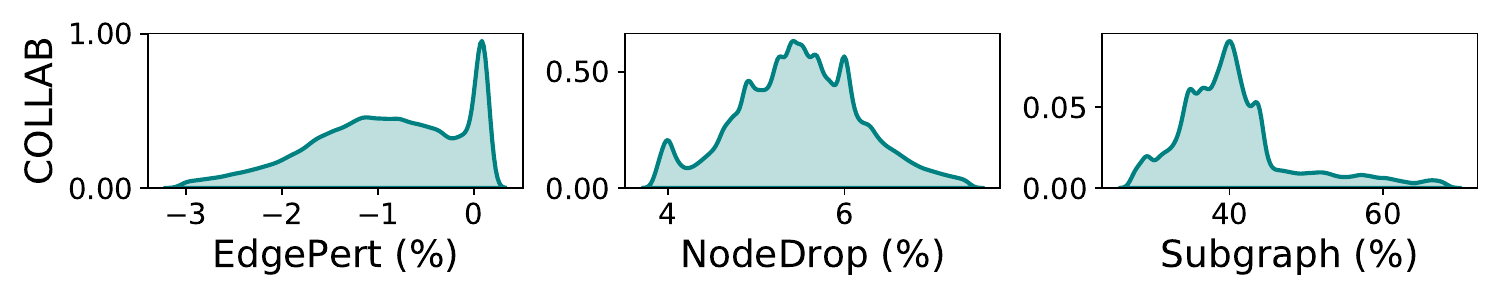}
  \end{subfigure}
  \\
  \begin{subfigure}{\linewidth}
    \centering
    \includegraphics[width=\linewidth]{REDDIT-BINARY_aug.pdf}
  \end{subfigure}
  \\
  \begin{subfigure}{\linewidth}
    \centering
    \includegraphics[width=\linewidth]{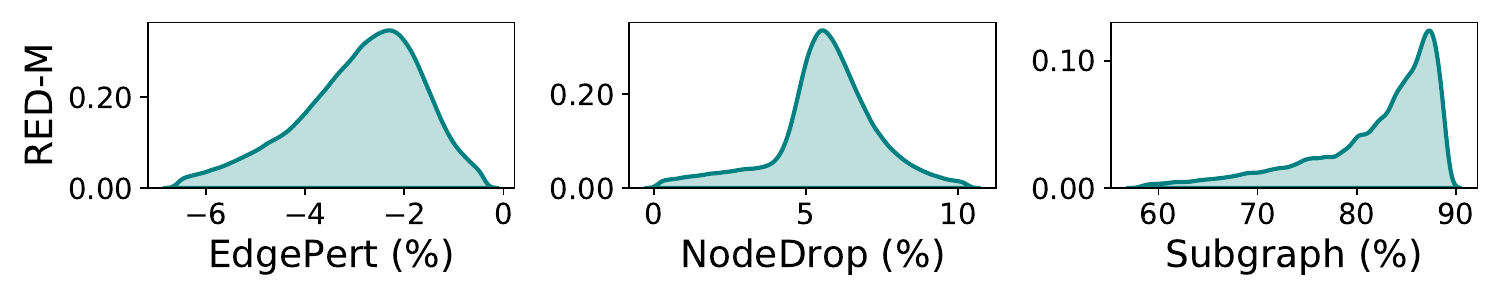}
  \end{subfigure}
  \\
  \begin{subfigure}{\linewidth}
    \centering
    \includegraphics[width=\linewidth]{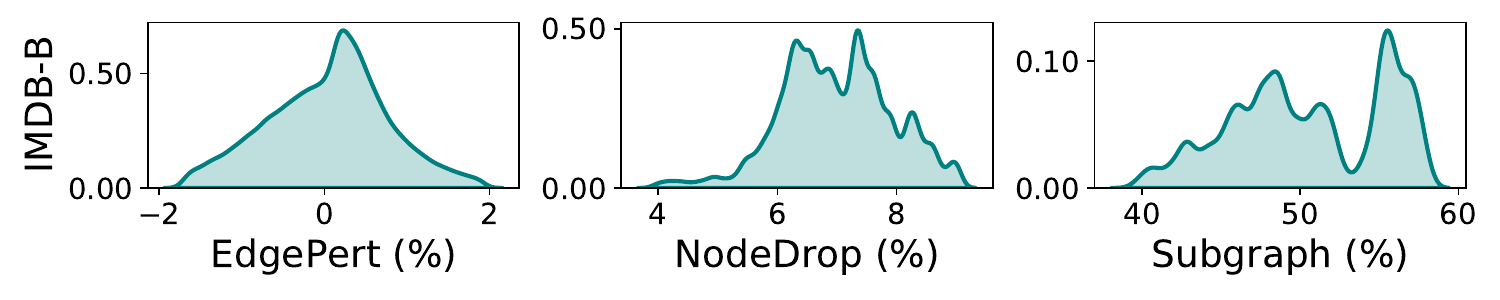}
  \end{subfigure}
  \caption{\textbf{Quantification of structural damage from data augmentation.} Percent change in structural entropy after data augmentation (i.e., Edge perturbation, Node dropping, and Subgraph with 20\% strength from GraphCL).} 
  \label{fig:quantification_structural_damage_all}
\end{figure}

\noindent \textbf{Data Augmentations on Graphs.}
Follow the data augmentations in GraphCL~\cite{you2020graph}, we adopt three types of general data augmentations for graph-structured data:
\begin{itemize}
  \item \textbf{Node dropping.} Given the graph $G$, node dropping will randomly discard certain portion of vertices along with their connections. The underlying prior enforced by it is that missing part of vertices does not affect the semantic meaning of $G$. Each node’s dropping probability follows a default i.i.d. uniform distribution (or any other distribution).
  
  \item \textbf{Edge perturbation.} It will perturb the connectivities in $G$ through randomly adding or dropping certain ratio of edges. It implies that the semantic meaning of $G$ has certain robustness to the edge connectivity pattern variances. We also follow an i.i.d. uniform distribution to drop each edge.

  \item \textbf{Subgraph.} This one samples a subgraph from $G$ using random walk. It assumes that the semantics of $G$ can be much preserved in its (partial) local structure.
\end{itemize}

The quantitative illustrations of structural damage caused by three data augmentation rules on eight datasets are shown in Figure~\ref{fig:quantification_structural_damage_all}.
As can be seen, the effect of structural damage varies with the augmentation rules.
Specifically, node dropping and subgraph lead to different degrees of structural damage, and the information loss caused by subgraph is the largest and generally over 50\%. 
Besides the simple information loss, the structure damage composition of edge perturbation is more complex; put differently, edge perturbation even introduces external data noise with the additional edges, which further interferes with the model from learning the actual structural information.

\begin{table}[!hp]
\centering
\caption{Statistics for datasets of diverse nature from the benchmark TUDataset.}
\label{tab:data_stat}
\resizebox{\linewidth}{!}{%
\begin{tabular}{l|cccc}
\hline \hline
Dataset & \#Graphs & \#Classes & Avg. \#Nodes & Avg. \#Edges \\ \hline \hline
\multicolumn{5}{c}{Social Networks} \\ \hline
COLLAB & 5,000 & 3 & 74.49 & 2457.78 \\
REDDIT-BINARY & 2,000 & 2 & 429.63 & 497.75 \\
REDDIT-MULTI-5K & 4,999 & 5 & 508.52 & 594.87 \\
IMDB-BINARY & 1,000 & 2 & 19.77 & 96.53 \\
\hline
\multicolumn{5}{c}{Small Molecules} \\ \hline
NCI1 & 4,110 & 2 & 29.87 & 32.30 \\
MUTAG & 188 & 2 & 17.93 & 19.79 \\ \hline
\multicolumn{5}{c}{Bioinformatics} \\ \hline
PROTEINS & 1,113 & 2 & 39.06 & 72.82 \\
DD & 1,178 & 2 & 284.32 & 715.66 \\ \hline \hline
\end{tabular}
}
\end{table}

\section{Summary of Datasets}
\label{sec:data_summary}
\subsection{Datasets for Unsupervised Learning}
A wide variety of datasets from different domains for a range of graph property prediction tasks are used for our experiments. Here, we present detailed descriptions of the 8 benchmarks utilized in this paper. Table~\ref{tab:data_stat} shows statistics for datasets.

\begin{table*}[!th]
\centering
\caption{Datasets statistics summary.}
\label{tab:data-stat-molecule}
\begin{tabular}{l|c|c|cccc}
\hline \hline
Dataset & Category & Utilization & \#Tasks & \#Graphs  & Avg.Node & Avg.Degree \\ \hline \hline
ZINC15  & Biochemical Molecules & Pre-Training & & 2,000,000 & 26.63    & 57.72      \\ \hline
BBBP    & Biochemical Molecules & Finetuning & 1       & 2,039     & 24.06    & 51.90      \\
Tox21   & Biochemical Molecules & Finetuning & 12      & 7,831     & 18.57    & 38.58      \\
ToxCast & Biochemical Molecules & Finetuning & 617     & 8,576     & 18.78    & 38.52      \\
SIDER   & Biochemical Molecules & Finetuning & 27      & 1,427     & 33.64    & 70.71      \\
ClinTox & Biochemical Molecules & Finetuning & 2       & 1,477     & 26.15    & 55.76      \\
MUV     & Biochemical Molecules & Finetuning & 17      & 93,087    & 24.23    & 52.55      \\
HIV     & Biochemical Molecules & Finetuning & 1       & 41,127    & 25.51    & 54.93      \\
BACE    & Biochemical Molecules & Finetuning & 1       & 1,513     & 34.08    & 73.71      \\ \hline \hline
\end{tabular}%
\end{table*}

\begin{figure*}[!ht]
  \centering
  \begin{subfigure}{\linewidth}
    \centering
    \includegraphics[width=\linewidth]{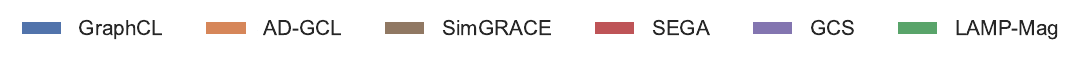}
    \vspace{-22pt}
  \end{subfigure} \\
  \begin{subfigure}{\linewidth}
    \centering
    \includegraphics[width=\linewidth]{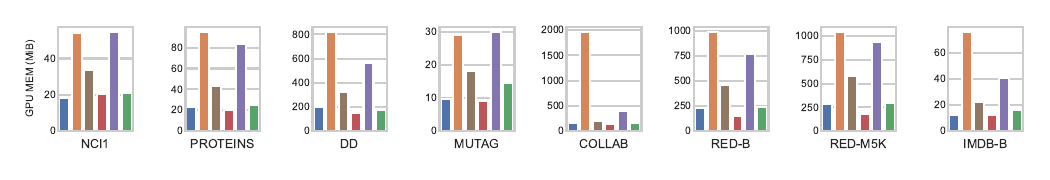}
    \vspace{-30pt}
    \caption{Training GPU Memory.}
    \label{fig:train_mem}
  \end{subfigure} \\
  \begin{subfigure}{\linewidth}
    \centering
    \includegraphics[width=\linewidth]{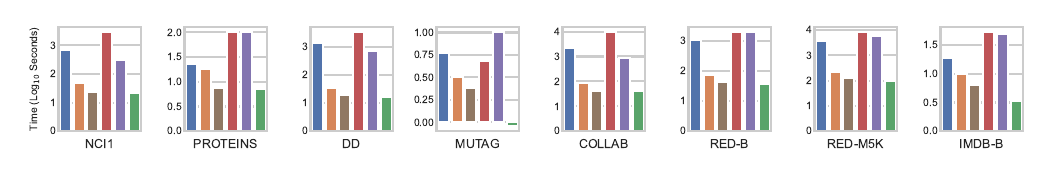}
    \vspace{-30pt}
    \caption{Training Time.}
    \label{fig:train_time}
  \end{subfigure} \\
  \begin{subfigure}{\linewidth}
    \centering
    \includegraphics[width=\linewidth]{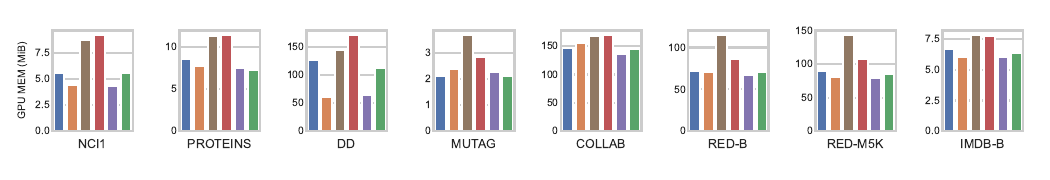}
    \vspace{-30pt}
    \caption{Inference GPU Memory.}
    \label{fig:inf_mem}
  \end{subfigure} \\
  \begin{subfigure}{\linewidth}
    \centering
    \includegraphics[width=\linewidth]{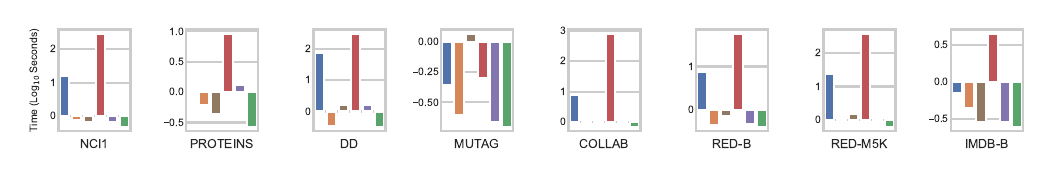}
    \vspace{-30pt}
    \caption{Inference Time.}
    \label{fig:inf_time}
  \end{subfigure} \\
  \caption{\textbf{Computation Cost.}}
  \label{fig:comp_cost}
\end{figure*}

\paragraph{Social Network Datasets.} IMDB-BINARY is derived from the collaboration of a movie set. In this dataset, every graph consists of actors or actresses, and each edge between two nodes represents their cooperation in a certain movie. Each graph is derived from a prespecified movie, and its label corresponds to the genre of this movie. Similarly, COLLAB is also a collaboration dataset but from a scientific realm, which includes three public collaboration datasets (i.e., Astro Physics, High Energy Physics and Condensed Matter Physics). Many researchers from each field form various ego networks for the graphs in this benchmark. The label of each graph is the research field to which the nodes belong. REDDIT-BINARY and REDDIT-MULTI-5K are balanced datasets, where each graph corresponds to an online discussion thread and nodes correspond to users. An edge is drawn between two nodes if at least one of them responds to another's comment. The task is to classify each graph into the community or subreddit to which it belongs.

\begin{figure}[!tp]
  \centering
  \begin{subfigure}{0.8\linewidth}
    \centering
    \includegraphics[width=0.4\linewidth]{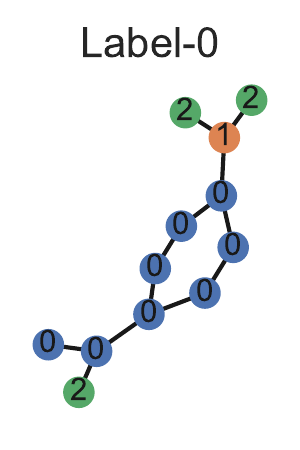} 
    \includegraphics[width=0.4\linewidth]{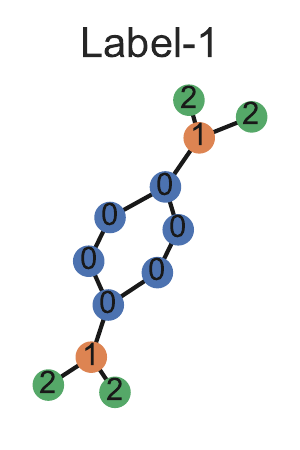}
    \caption{Negative Samples.}
    \label{fig:neg_sample}
  \end{subfigure} \\
  \begin{subfigure}{0.48\linewidth}
    \centering
    \includegraphics[width=\linewidth]{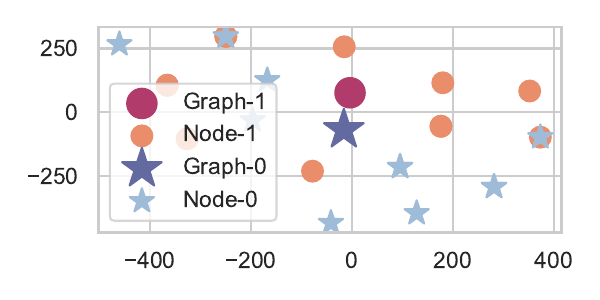}
    \caption{TSNE w/o $\mathcal{L}_{LocalC}$.}
    \label{fig:neg_tsne_wo}
  \end{subfigure} 
  \begin{subfigure}{0.48\linewidth}
    \centering
    \includegraphics[width=\linewidth]{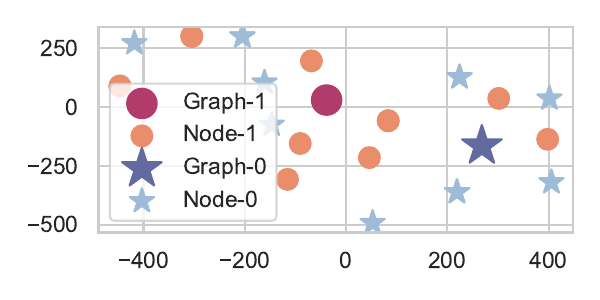}
    \caption{TSNE w/ $\mathcal{L}_{LocalC}$.}
    \label{fig:neg_tsne}
  \end{subfigure}
  \caption{Hard Negative Samples in MUTAG.}
  \label{fig:hard_neg}
\end{figure}

\begin{figure*}[!h]
\centering
  \includegraphics[width=\linewidth]{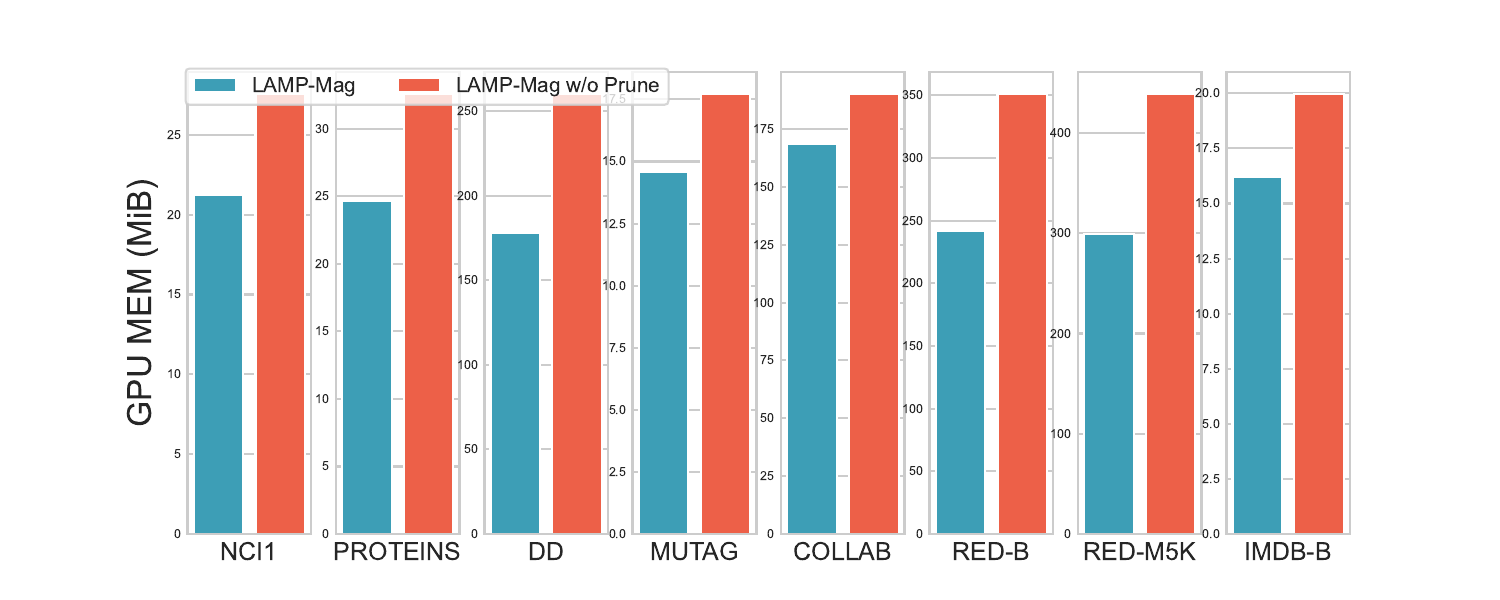}
  \caption{GPU memory usage with and without model pruning.}
 \label{fig:prune_comp}
\end{figure*}

\paragraph{Small Molecules.} NCI1 is a dataset made publicly available by the National Cancer Institute (NCI) and is a subset of balanced datasets containing chemical compounds screened for their ability to suppress or inhibit the growth of a panel of human tumor cell lines; this dataset possesses 37 discrete labels. 
MUTAG has seven kinds of graphs that are derived from 188 mutagenic aromatic and heteroaromatic nitro compounds. PTC includes 19 discrete labels and reports the carcinogenicity of 344 chemical compounds for male and female rats.

\paragraph{Bioinformatic Datasets.} DD contains graphs of protein structures. A node represents an amino acid and edges are constructed if the distance of two nodes is less than $6\AA$. A label denotes whether a protein is an enzyme or non-enzyme. 
PROTEINS is a dataset where the nodes are secondary structure elements (SSEs), and there is an edge between two nodes if they are neighbors in the given amino acid sequence or in 3D space. The dataset has 3 discrete labels, representing helixes, sheets or turns.

\subsection{Details of Molecular Datasets}
\paragraph{Input graph representation.} For simplicity, we use a minimal set of node and bond features that unambiguously describe the two-dimensional structure of molecules. We use RDKit \cite{landrum2013rdkit} to obtain these features.

\begin{itemize}
    \item Node features:
    \begin{itemize}
        \item Atom number: [1, 118]
        \item Chirality tag: \{unspecified, tetrahedral cw, tetrahedral ccw, other\}
    \end{itemize}
    \item Edge features:
    \begin{itemize}
        \item Bond type: \{single, double, triple, aromatic\}
        \item Bond direction: \{–, endupright, enddownright\}
    \end{itemize}
\end{itemize}

\paragraph{Downstream task datasets.} 8 graph classification datasets from MoleculeNet \cite{wu2018moleculenet} are used to evaluate model performance.

\begin{itemize}
    \item BBBP \cite{martins2012bayesian}. Blood-brain barrier penetration (membrane permeability), involves records of whether a compound carries the permeability property of penetrating the blood-brain barrier.
    \item Tox21 \cite{Tox21}. Toxicity data on 12 biological targets, which has been used in the 2014 Tox21 Data Challenge and includes nuclear receptors and stress response pathways.
    \item ToxCast \cite{richard2016toxcast}. Toxicology measurements based on over 600 in vitro high-throughput screenings.
    \item SIDER \cite{kuhn2016sider}. Database of marketed drugs and adverse drug reactions (ADR), grouped into 27 system organ classes and also known as the Side Effect Resource.
    \item ClinTox \cite{novick2013sweetlead,gayvert2016data}. Qualitative data classifying drugs approved by the FDA and those that have failed clinical trials for toxicity reasons.
    \item MUV \cite{gardiner2011effectiveness}. Subset of PubChem BioAssay by applying a refined nearest neighbor analysis, designed for validation of virtual screening techniques.
    \item HIV \cite{HIV}. Experimentally measured abilities to inhibit HIV replication.
    \item BACE \cite{subramanian2016computational}. Qualitative binding results for a set of inhibitors of human $\beta$-secretase 1.
\end{itemize}

\paragraph{Details of Dataset Splitting}
For molecular prediction tasks, following \cite{ramsundar2019deep}, we cluster molecules by scaffold (molecular graph substructure) \cite{bemis1996properties}, and recombine the clusters by placing the most common scaffolds in the training set, producing validation and test sets that contain structurally different molecules. Prior work has shown that this scaffold split provides a more realistic estimate of model performance in prospective evaluation compared to random split \cite{chen2012comparison,sheridan2013time}. The split for train/validation/test sets is 80\%:10\%:10\%.

\begin{figure}[!th]
\centering
  \includegraphics[width=0.8\linewidth]{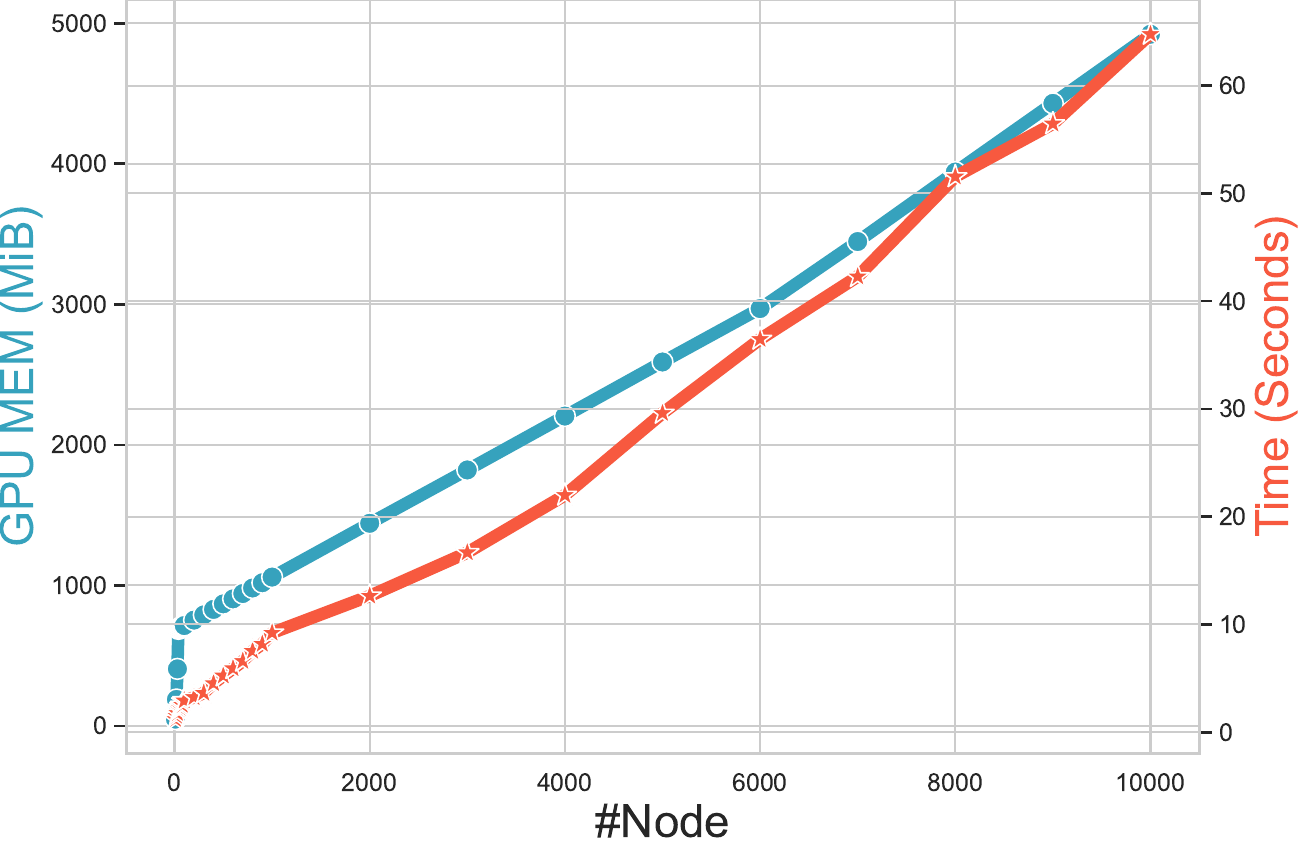}
  \caption{Scalability.}
  \label{fig:scalability}
\end{figure}

\section{Detailed Experiment Setup}
\label{sec:exp_setup}
\subsection{Settings for Unsupervised Learning}
\noindent \textbf{Datasets.} Eight benchmarks are adopted from TUDataset~\cite{morris2020tudataset} and summarized in Table~\ref{tab:data_stat}, including IMDB-BINARY, REDDIT-MULTI-5K, NCI1, MUTAG, PROTEINS, DD, REDDIT-BINARY, and COLLAB.

\noindent \textbf{Configuration.} Hidden dimension is chosen from $\{32,64\}$, and batch size is chosen from $\{32,128\}$. An Adam optimizer \cite{kingma2015adam} is employed to minimize the contrastive lose with $\{0.01, 0.005, 0.001\}$ learning rate.

\noindent \textbf{Learning protocols.} In unsupervised representation learning \cite{sun2020infograph}, all data is used for model pre-training and a non-linear SVM is adopted as classifier to perform to perform 10-fold cross-validation on learned graph embeddings.
For graph representation learning, models are trained 20 epochs and tested every 10 epochs. 
The 10-fold evaluation are performed 5 times in total with different random seeds as~\cite{sun2020infograph}. At last, we report the average accuracy and standard deviation (\%). 

\subsection{Setting for Transfer Learning}
\noindent \textbf{Pre-training dataset.}
ZINC15 \cite{sterling2015zinc} dataset is adopted for pre-training. In particular, a subset with two million unlabeled molecular graphs is sampled from the ZINC15.

\noindent \textbf{Pre-training details.} 
In the graph encoder setting in \cite{hu2020strategies}, GIN \cite{xu2019powerful} with five convolutional layers is adopted for message passing. In particular, the hidden dimension is fixed to 300 across all layers and a pooling readout function that averages graph nodes is hired for NT-Xent loss calculation with the scale parameter $\tau = 0.1$. The hidden representations at the last layer are injected into the average pooling function.
An Adam optimizer \cite{kingma2015adam} is employed to minimize the integrated losses produced by the 5-layer GIN encoder. All training processes will run 100 epochs with a batch size of 256.

\noindent \textbf{Fine-tuning dataset.} 
We employ the eight ubiquitous benchmarks from the MoleculeNet dataset \cite{wu2018moleculenet} as the downstream experiments. These benchmarks include a variety of molecular tasks like physical chemistry, quantum mechanics, physiology, and biophysics. 
For dataset split, the scaffold split scheme \cite{chen2012comparison} is adopted for train/validation/test set generation. 
Table~\ref{tab:data-stat-molecule} summarizes the basic characteristics of the datasets.

\noindent \textbf{Fine-tuning details.} 
For downstream tasks, a linear layer is stacked after the pre-trained graph encoders for final property prediction. The downstream model still employs the Adam optimizer for 100 epochs of fine-tuning. All experiments on each dataset are performed for ten runs with different seeds, and the results are the averaged ROC-AUC scores (\%) $\pm$ standard deviations. 
The alternatives of learning rate in pre-training and fine-tuning phases are \{0.0001, 0.001, 0.01\}.
To be in line with \cite{you2020graph}, the epochs for pre-training range from 20 to 100 with a step of 20 epochs.

\section{Computation Cost}
Besides the superior performance, here, we further analyze the computation efficiency of our methods, including the GPU memory consumption and time in training and inference phase.
As depicted in Figure~\ref{fig:comp_cost}, LAMP-Mag (pruning ratio of 50\%) gives the shortest training and inference time, while maintaining a comparable GPU memory cost. 
Moreover, we show the GPU memory cost of LAMP-Mag with and without pruning in Figure~\ref{fig:prune_comp}, the magnitude pruning effectively reduce the computation overhead, especially for large graphs (i.e., DD and RED-M5K). Detailed discussion will be given in the new revision.

\section{Hard Negative Samples}
As shown in Figure~\ref{fig:neg_sample}, the two samples from MUTAG dataset have the same structure but different node labels. As shown in Figure~\ref{fig:neg_tsne_wo} and \ref{fig:neg_tsne}, our model successfully enlarge the distance of the two samples with the introduction of local contrastive loss. 

\section{Scalability}
We evaluate the scalability of LAMP-Mag on Erdos-Renyi graphs with $|E| = 2|V|$.
In Figure~\ref{fig:scalability}, LAMP-MAG demonstrates high memory efficiency that nearly 5GiB GPU memory is needed to train a batch (batch size of 128) of graphs with 10,000 nodes.









\end{document}